\documentclass[twoside]{article}

\usepackage[accepted]{aistats2025}

\usepackage{tikz}
\usepackage{tikz-3dplot}
\usetikzlibrary{calc}
\usetikzlibrary{decorations.pathreplacing}

\usepackage[inkscapelatex=false]{svg}

\usepackage{multicol}
\usepackage{hyperref}
\usepackage{url}

\usepackage{times}
\usepackage{latexsym}
\usepackage{rotating}
\usepackage{tabularray}

\usepackage{svg}

\usepackage{tabularx}
\usepackage{fancyvrb}
\usepackage{cprotectinside}
\DefineVerbatimEnvironment{Code}{BVerbatim}{baseline=t}

\usepackage{minitoc}

\usepackage[T1]{fontenc}

\usepackage[utf8]{inputenc}

\usepackage{microtype}

\usepackage{inconsolata}

\usepackage{graphicx}

\usepackage{hyperref}       
\usepackage{url}            
\usepackage{booktabs}       
\usepackage{amsfonts}       
\usepackage{nicefrac}       
\usepackage{microtype}      
\usepackage{xcolor}         

\usepackage{float}

\usepackage[round]{natbib}

\usepackage{tabularx}
\usepackage{bbm}
\usepackage{makecell}
\usepackage{booktabs}
\usepackage{multirow}
\usepackage{amsmath}
\usepackage{rotating}
\usepackage{enumitem}
\usepackage{xcolor}
\usepackage{bbding}
\usepackage{algorithm}
\usepackage{algorithmic}
\usepackage{amsmath}
\usepackage{amssymb}
\usepackage{mathtools}
\usepackage{amsthm,bm}
\usepackage{ mathrsfs }
\usepackage[capitalize,noabbrev]{cleveref}
\usepackage{lipsum}
\usepackage{subcaption}
\usepackage{caption}

\usepackage{rotating}
\usepackage{tabularray}
\usepackage{color}
\usepackage[normalem]{ulem}
\usepackage{rotating}
\usepackage{tabularray}

\usepackage{enumitem}
\usepackage{multirow}
\usepackage{xcolor}
\usepackage{hyperref}
\usepackage{microtype}
\usepackage{graphicx}
\usepackage{booktabs} 
\usepackage{diagbox}
\usepackage{mathtools}
\usepackage{wrapfig}
\usepackage{comment}

\usepackage{color}
\usepackage{tabularray}
\usepackage{makecell}
\usepackage{multirow}
\usepackage{booktabs}

\theoremstyle{plain}
\newtheorem{theorem}{Theorem}[section]

\newtheorem{proposition}{Proposition}
\newtheorem{lemma}{Lemma}

\theoremstyle{definition}
\newtheorem{definition}{Definition}

\theoremstyle{remark}
\newtheorem{remark}{Remark}

\newcommand{\piref}{\pi_\text{ref}}
\newcommand{\pisft}{\pi^\text{SFT}} %

\def\b0{\mathbf{0}}
\def\b1{\mathbf{1}}

\def\cD{\mathcal{D}}

\def\cX{\mathcal{X}}

\def\mR{\mathbb{R}}

\def\EE{\mathbb{E}}
\def\mP{\mathbb{P}}

\DeclareMathOperator*{\argmax}{arg\,max}

\newcommand{\mbrkt}[1]{\left(#1\right)}

 \newcommand{\black}[1]{{\color{black}{#1}}}

\newcommand{\cM}{\mathcal{M}}      \newcommand{\ct}{\mathcal{T}}  \newcommand{\cv}{\mathcal{V}}  \newcommand{\cx}{\mathcal{X}} \newcommand{\cy}{\mathcal{Y}} 

\newcommand{\eps}{\varepsilon}

\DeclarePairedDelimiter\abs{\lvert}{\rvert}%
\DeclarePairedDelimiter\norm{\lVert}{\rVert}%

\makeatletter
\let\oldabs\abs
\def\abs{\@ifstar{\oldabs}{\oldabs*}}
\let\oldnorm\norm
\def\norm{\@ifstar{\oldnorm}{\oldnorm*}}
\makeatother

\begin{document}

\runningtitle{Exposing Privacy Gaps in LLM Alignment}

%
\runningauthor{Feng et. al.}

\twocolumn[

\aistatstitle{Exposing Privacy Gaps: Membership Inference Attack \\ on Preference Data for LLM Alignment}

\aistatsauthor{ Qizhang Feng$^*$  \And Siva Rajesh Kasa$^*$ \And  Santhosh Kasa \AND Hyokun Yun \And Choon Hui Teo \And Sravan Bodapati }

\aistatsaddress{Amazon Inc.}

%
%

]


\begin{abstract}
	Large Language Models (LLMs) have seen widespread adoption due to their remarkable natural language capabilities. However, when deploying them in real-world settings, it is important to align LLMs to generate texts according to acceptable human standards. Methods such as Proximal Policy Optimization (PPO) and Direct Preference Optimization (DPO) have enabled significant progress in refining LLMs using human preference data. However, the privacy concerns inherent in utilizing such preference data have yet to be adequately studied. In this paper, we investigate the vulnerability of LLMs aligned using two widely used methods - DPO and PPO - to membership inference attacks (MIAs). Our study has two main contributions: first, we theoretically motivate that DPO models are more vulnerable to MIA compared to PPO models; second, we introduce a novel reference-based attack framework specifically for analyzing preference data called PREMIA (\uline{Pre}ference data \uline{MIA}). Using PREMIA and existing baselines we empirically show that DPO models have a relatively heightened vulnerability towards MIA. 
\end{abstract}

%

\def\thefootnote{*}\footnotetext{These authors contributed equally to this work. \\ Contact: \{qzf,kasasiva\}@amazon.com}

\section{INTRODUCTION}
Large language models (LLMs) have seen a surge in their adoption in the recent past due to their remarkable capabilities on a wide range of natural language processing (NLP) tasks such as question answering, code generation, etc \citep{zhao2023survey}. When deployed in real-world scenarios, it is important to align LLMs to human preferences. Techniques such as Proximal Policy Optimization (PPO) and Direct Preference Optimization (DPO) play a key role in aligning LLMs with human ethical standards by leveraging human-derived preference data ~\citep{christiano2017deep,rafailov2024direct,yang2024harnessing}. Although these approaches improve the alignment of models with human values, they are fraught with privacy concerns because of their use of human-generated data. In this work, we investigate the Membership Inference Attack (MIA), a widely-studied vulnerability that attempts to determine whether if specific data points are part of the model's preference dataset. The study of MIA highlights vulnerabilities in a variety of machine learning paradigms, including several recent studies that specifically focus on LLMs \citep{fu2023practical, shi2024detecting}.
Although existing research on MIA in the context of LLMs highlights the need to evaluate and address the need for privacy concerns, the unique challenges posed by alignment methods such as the PPO and DPO approaches (where preference data directly influences model behavior) remain to be explored. Traditional MIA frameworks fall short when applied to the complex, context-dependent optimization procedures used in LLM alignment. In this paper, we theoretically and empirically show that reward-model-free approaches such as DPO are more susceptible to MIA, as compared to the reward-model-based approaches such as PPO. We also introduce a novel reference-model based MIA framework that is specifically tailored towards preference data and LLM alignment, providing a more precise analysis tool that can better highlight these vulnerabilities. Our contributions to this field are twofold:
\begin{itemize}
		\item \textbf{Comparative Vulnerability Assessment of DPO and PPO Models:} 
		First, we theoretically motivate that despite being asymptotically equivalent in the optimization objective, DPO tends to overfit on the preference data vis à vis PPO. Next we derive a generic lower bound on the Bayes optimal membership on algorithms that tend to overfit, which shows why DPO models are more vulnerable to MIA compared to PPO models..
	\item \textbf{Introduction of a Novel Reference-based Attack Framework:} Following up on the theoretical overfitting results, we empirically demonstrate the differential susceptibility of DPO vs PPO to MIA on two real world datasets. We propose PREMIA, a comprehensive optimistic attack framework tailored to assess the vulnerability of preference data to MIA, providing a practical upper bound to quantify the extent of attack effectiveness. Through PREMIA, along with existing MIA frameworks, our experiments highlight the complex interplay between model sizes and task difficulty in LLM alignment. 

\end{itemize}

\section{PRELIMINARIES}
This section introduces the notations and background concepts required for the rest of the paper. We begin by defining the frameworks of PPO and DPO, followed by an overview of MIAs.

\subsection{Model Alignment}
Model alignment ensures LLMs adhere to human values and ethics by adjusting their outputs to match human preferences~\citep{hendrycks2021unsolved,ouyang2022training}. Such alignment is critical for creating AI systems that act in ways that benefit humans and reduce the risks associated with improper alignment. Among the various model alignment techniques, PPO and DPO are some of the widely used approaches \cite{xu2024dpo}.

\subsubsection{Proximal Policy Optimization (PPO)}
\citet{stiennon2020learning} and \citet{bai2022training} illustrate Reinforcement Learning from Human Feedback (RLHF) that integrates human feedback into the alignment of pre-trained Language Models (LMs), encompassing three phases: Supervised Fine-Tuning (SFT), Preference Sampling with Reward Learning, and Reinforcement Learning (RL) through PPO.

\textbf{SFT} begins the process by fine-tuning a pre-trained LM on task-specific data to obtain a model $\pisft$, enhancing the LLM's performance on the task at hand.

\textbf{Preference Data Collection} involves gathering a set of preference data pairs $(x, y_w, y_l)$, where $x$ is a prompt and $y_w$, $y_l$ are two different responses. Here, $y_w$ is the response preferred by human evaluators over $y_l$ for the given context $x$.

\textbf{Reward Modeling Phase} uses the preference pairs to train the reward model $r_{\phi}(x, y)$, where $\phi$ represents the trainable parameters. The trainable model can be a classification header layer attached to the base model or a separate model. The Bradley-Terry (BT) model is commonly used to represent the probability that one response is better than another:
\begin{align} 
 \label{eq:RW_model}
\max_{\phi}	-\mathbb{E}_{(x, y_w, y_l) \sim \mathcal{D}}\left[\log \sigma(r_{\phi}(x, y_w) - r_{\phi}(x, y_l))\right],
\end{align}
where $r_{\phi}(x, y)$ models the likelihood of preferring $y_w$ to $y_l$ given the prompt $x$, and $\mathcal{D}$ denotes the dataset of preference pairs. This loss function measures the accuracy of the reward model in predicting human preferences.

\textbf{RL Fine-Tuning Phase} then fine-tunes the LM further using the learned reward function, striving to align model outputs with human preferences while maintaining generative diversity:
\begin{align}
	\max_{\pi_{\theta}}  \mathbb{E}_{x \sim \mathcal{D}, y \sim \pi_{\theta}(y | x)}[r_{\phi}(x, y)] - \beta \mathbb{D}_{\textrm{KL}}[\pi_{\theta}(y | x) || \pisft(y | x)], \label{eq:PPO}
\end{align}
balancing fidelity to human feedback with the preservation of the model's original capabilities. Here, $\pi_{\theta}$ represents the policy of the language model parameterized by $\theta$, the trainable parameters. The optimization in Equation \ref{eq:PPO} is carried out using Proximal Policy Optimization (PPO) method \citep{schulman2017proximal}.

\subsubsection{Direct Preference Optimization (DPO)}
DPO offers a simplified approach towards aligning language models by directly leveraging preference data, bypassing the explicit reward model construction typically associated with PPO methodologies \citep{rafailov2024direct}. This method reformulates the two-step optimization procedure in Equations \ref{eq:RW_model} and \ref{eq:PPO} into a single optimization problem that simultaneously optimizes the policy and encodes an implicit reward mechanism based on the preference data as follows: 

\citet{rafailov2024direct} show that the optimal policy that maximizes Equation \ref{eq:PPO} can be written as 
\begin{align}
	\pi^*(y \mid x)=\frac{1}{Z(x)} \pi_{\mathrm{ref}}(y \mid x) \exp \left(\frac{1}{\beta} r_{\phi}(x, y)\right) \label{eq:optml_ppo_policy}
\end{align}
where $Z(x)$ is a partition function which depends on the prompt $x$ \cite{rafailov2024direct,xu2024dpo}. Rearranging Equation \ref{eq:optml_ppo_policy} gives the corresponding reward as follows:
\vspace{-\baselineskip}
\begin{align}
	r_{\phi}(x, y) = \beta \log \frac{\pi_{\theta}(y_w\mid x)}{\piref(y_w\mid x)} + C(x),  \label{eq:optml_ppo_reward}
\end{align}
where $C: \cx \to \mR$ is a scalar function. Here, $\piref$ refers to a reference model which is typically chosen as the SFT model $\pisft$. 
\begin{align}
&\max_{\pi_{\theta}}  -\mathbb{E}_{(x, y_w, y_l) \sim  \mathcal{D}}   \label{eq:DPO} \\
	&\biggl[\log \sigma \biggl(\beta \log \frac{\pi_{\theta}(y_w\mid x)}{\piref(y_w\mid x)} - \beta \log \frac{\pi_{\theta}(y_l\mid x)}{\piref(y_l\mid x)} \biggr) \biggr]. \nonumber
\end{align}
This optimization method is often preferred over PPO because it simplifies training by optimizing directly on the preference data, which improves computational efficiency and is easier to implement ~\citep{rafailov2024direct,xu2024dpo}. Note that in PPO (equation \ref{eq:PPO}), contrary to DPO (equation \ref{eq:DPO}), the final model being optimized is not directly aligned using the data $\mathcal{D}$. This is the key intuition behind why PPO-aligned models are less susceptible to privacy threats compared to their DPO counterparts.
Next, we state a result about the implicit reward induced by $\pi_{DPO}$. 

\begin{lemma}\label{lemma:dpo_implicit_reward}
	Let $\pi_{DPO}$ be the policy obtained by optimizing Equation \ref{eq:DPO}. Then, a) the corresponding implicit rewards on the preference pairs that optimize the BT model loss in Equation \ref{eq:RW_model} are given by $r_d(x,y_w) = \beta \log \frac{\pi_{DPO}(y_w|x)}{\pi_{ref}(y_w|x)}$ and $r_d(x,y_l) = \beta \log \frac{\pi_{DPO}(y_l|x)}{\pi_{ref}(y_l|x)}$ and b) $\pi_{DPO}$ maximizes the objective $ \max_{\pi_{\theta}}  \mathbb{E}_{x \sim \mathcal{D}, y \sim \pi_{\theta}(\{y_l,y_w\} | x)}[r_d(x, y)] - \beta \mathbb{D}_{\textrm{KL}}[\pi_{\theta}(y | x) || \pi_{\text{ref}}(y | x)]$. 
\end{lemma}

\begin{proof}
	Claim a) directly follows by subsitiuting the values of $r_d(x,y_w)$ and $r_d(x,y_l)$ in Equation \ref{eq:RW_model}. Proof of Claim b) exactly follows the same steps as that of how Equation \ref{eq:optml_ppo_policy} is the optimal policy for the Objective \ref{eq:PPO} - given in \citet{rafailov2024direct}.
\end{proof}

\subsection{Membership Inference Attacks (MIA) on LLMs}
MIA poses a significant privacy risk in the context of LLMs, challenging the security of data used in training such models \citep{shokri2017membership,nasr2018comprehensive}. In LLMs, MIAs seek to determine whether specific data was part of the model's training set, exploiting the model's behavior or output nuances to infer data membership. These attacks are particularly concerning for models trained on vast datasets, where inadvertently revealing individual data points could lead to privacy breaches.

The effectiveness of an MIA against LLMs is quantified by a score function $\mathcal{M}$, mapping input samples and the level of access to a real-valued score indicating the likelihood of membership. For a given threshold $\tau$, an input $x$ is classified as a training set member if $\mathcal{M}(x, \text{Access}(\theta)) \geq \tau$.
\begin{equation}
	\mathcal{M}:\mathcal{X}\times \text{Access}(\Theta) \to\mathbb{R}.
\end{equation}
Broadly speaking, MIAs on LLMs can be divided into two categories - Reference-based attacks and Reference-free attacks \citep{fu2023practical}. The effectiveness of Reference-based attacks relies on the assumption that training records exhibit a higher probability of being sampled compared to non-training records. By comparing the output probabilities of the target model against those of the reference model, attackers can infer membership status more reliably. Reference-free attacks don't hinge on any such above mentioned assumption and solely rely on the model's tendency to overfit, if any. Thus, they are less effective compared to their Reference-based counterparts. 
Mattern et. al. (2023) propose a reference-free MIA known as the Neighbour attack specifically for finetuned LLMs; this approach relies on generating synthetic semantically equivalent samples known as Neighbours, given a target point from the finetuning corpus and comparing the model score of the neighboring points to that of the target point. The intuition is that if the model score of the target point is similar to that of the crafted neighbours, then the target point, along with the crafted neighbors are plausible points from the same distribution and the target point is not part of the training set. However, if the model score of the target point is significantly higher compared to the crafted neighbors, then it is more likely that the target point is part of the finetuning data. Zhang et. al. (2024) propose a reference-free MIA known as Divergence-based Calibration Method (DC-PDD) for detecting MIA vulnerabilities for pretraining data of LLMs. DC-PDD relies on publicly available pretraining corpus to construct an within-document token frequency distribution and comparing its divergence against the token probability distribution to derive a detection score. Xie et. al. (2024) propose a reference-free MIA known as Relative Conditional Log Likelihood (ReCALL) for detecting MIA vulnerabilities associated with pretraining data. The idea is to prefix non-member data, which can be obtained using recent data beyond the model training cut-off or generated synthetically, to target points and showing a differential effect on member texts from pretrained corpus vs non-member texts. Note that none of the existing MIAs are designed to factor in the tuple
structure of preference data. Further more, the MIAs designed for pretraining data typically have not been effective as the volume of pretraining data is orders of magnitude higher compared to finetuning or RLHF alignment.
\citet{duan2024membership} conducted an exhaustive experimentation of various MIA attacks (both reference-based and free) on different pretrained LLMs, and observed that most MIAs barely outperform random guessing. This was primarily attributed to two main reasons: a) the large volume of natural language text on which LLMs are pretrained makes the member domains indistinguishable to attacks and b) near-one epoch training prevents over-memorization of pretraining data. \citet{maini2024llm} also acknowledge a similar issue with respect to existing MIAs on pretrained LLMs and propose novel complementary attack framework known as Dataset Inference Attack. Research on MIAs targeting LLMs underscores the need for robust privacy-preserving techniques to safeguard training data, with implications for the development and deployment of secure, trustworthy AI systems \citep{carlini2020extracting}.

\subsection{Problem Statement}
Current research on MIAs has advanced understanding of risks in pre-trained text models, but gaps remain in applying MIAs to preference datasets in LLM alignment. This oversight poses substantial privacy risks, given the critical role of preference data in shaping LLM outputs.

Let $\mathcal{D} = \{(x_i, y_{wi}, y_{li})\}_{i=1}^{N}$ represent the preference dataset, where $x_i$ is a prompt, $y_{wi}$ is the preferred response, and $y_{li}$ is the less preferred response. The vulnerability of this preference data to MIAs requires a nuanced examination, which can be categorized into three distinct attack vectors:


%
%

\begin{itemize}
	\item \textbf{Attack against prompts and individual responses:} This attack determines whether a specific pair of prompt $x$ and response $y_w$  or $y_l$has been used in training, highlighting potential privacy breaches if such data can be identified:
	\item \textbf{Attack against the entire preference tuple:} This more comprehensive attack assesses whether the entire tuple $(x, y_w, y_l)$ can be traced back to the preference set. Note that traditional attack frameworks for SFT or pre-training have only focused on prompts and responses, and have not examined this privacy risk associated with tuples. 
\end{itemize}
This detailed breakdown elucidates the complex vulnerabilities associated with preference data in LLMs. By identifying these specific attack vectors, we aim to advance privacy-preserving mechanisms that safeguard the alignment process and ensure that models respect and protect individual privacy while adhering to human ethical standards.

\subsection{Theoretical Results}

In this section, using analysis similar to that of \cite{li2023policy}, we first show that DPO has a tendency to overfit on the preference dataset, compared to PPO. Using these results, we comment on the bayes optimal membership inference for both DPO and PPO.  We defer all the proofs to Appendix \ref{app:proofs}.

Let \( \mathcal{D}_{\text{pair}} = \left\{ (x_i, y_i) \right\}_{i=1}^n \) be the prompt-response dataset where each \( (x_i, y_i) \) is a data pair with \( x_i \) being a prompt sampled from the distribution \( \cX \) and \( y_i \) being a response sampled from the conditional distribution $\pi(.|x_i )$. \black{Note that it is possible to construct the preference dataset $\cD$ from this notation by repeating the same $x_i$ for both $y_{i_w}$ and $y_{i_l}$. }
Let $r(x,y)$ be the true reward for any data pair $(x,y)$ and let $\pi_{r}^{*}$ be the policy that maximizes the expected reward i.e. $$\pi_r^\star \leftarrow \arg\max_\pi \mathbb{E}_{x \sim \cX} \mathbb{E}_{y \sim \pi(.|x)} [r(x,y)].$$


In practice, we don't know the true reward model $r$. Neither can we compute the true expectations with respect to $\cX$ and $\pi(.|x)$. Let $\hat{r}_p$ and $\pi_{PPO}$ be the explicit reward and explicit policy estimated through PPO. Let $\hat{r}_d$ and $\pi_{DPO}$ be the implicit reward and explicit policy estimated through DPO. For ease of exposition, we ignore the regularization terms i.e. we assume $\beta =0$. Then the PPO and DPO objectives can be stated as follows \cite{li2023policy}:
\begin{align}
	\pi_{PPO} & \leftarrow \underset{\pi}{\operatorname{argmax}} \sum_{i=1}^n \EE_{y \sim \pi(.|x_i)} \widehat{r}_p\left(x_i, y\right) \label{eq:toy_ppo_eq}\\
	\pi_{DPO} & \leftarrow \underset{\pi}{\operatorname{argmax}} \sum_{i=1}^n \widehat{\EE}_{y \sim \hat{\pi}(.|x_i)} \widehat{r}_d\left(x_i, y\right) \label{eq:toy_dpo_eq}
\end{align}
where, $\hat{\EE}$ denotes the finite sample expectation of the response given a prompt.
\begin{remark}
	 Equation \ref{eq:toy_ppo_eq} follows from Equation \ref{eq:DPO} and Equation \ref{eq:toy_dpo_eq} follows from claim b) of Lemma \ref{lemma:dpo_implicit_reward}. Note that it is possible that the learned reward model could err for some $x$ i.e. $\hat{r}(x,y_l) > \hat{r}(x,y_w)$
\end{remark}

Following \cite{li2023policy}, we define three types of error:
\begin{enumerate}
	\item reward estimation error defined as $\eps_r := \sup_{\hat{r}(x,y)}\abs{\widehat{r}(x,y) - r(x,y)}$,
	\item prompt estimation error as $\eps_x :=  \abs{ \EE_x \EE_{y \sim \pi(\cdot|x)} [r(x,y)] - \hat{\EE}_x \EE_{y \sim \pi(\cdot|x)} [r(x,y)]}$, and
	\item the response distribution estimation error $\varepsilon_y := \sup_{\pi, r} \sup_x \abs{ \mathbb{E}_{y \sim \pi(\cdot|x)} [r(x,y)] - \hat{\mathbb{E}}_{y \sim \pi(\cdot|x)} [r(x,y)] }$
\end{enumerate}
Here, $\sup_{\pi, r} := \sup_\pi \sup_{r: \cX \times \cy}$; and  $\hat{\EE}_x$ and $\hat{\EE}_y$ denote the finite-sample estimations of expectation under the preference dataset's prompt and response distributions, respectively. 

Let $\pi_r^{*}$ be the optimal policy which maximizes the true reward over the preference dataset i.e. $\pi_r^{*} := \argmax_{\pi} \hat{\EE}_x\hat{\EE}_{y\sim \pi} [r(x,y)]$. Let $r(\pi)$ be the finite sample expectation true reward for any policy $\pi$ i.e. $r(\pi) := \hat{\EE}_x\hat{\EE}_{y\sim \pi} [r(x,y)]$  

\begin{proposition}\label{lemma:overfit_ours}
	$r(\pi_r^{*}) - r(\pi_{\text{PPO}}) \leq 2\eps_r + 2\eps_x$ ; $r(\pi_r^{*}) - r(\pi_{\text{DPO}}) \leq 2\eps_r $
\end{proposition}

If $\pi_r^{*}$ and $r(\pi)$ were defined on the population rather than the preference dataset i.e. $\pi_r^{*} := \argmax_{\pi} \EE_x \EE_{y\sim \pi} [r(x,y)]$ and $r(\pi) := \EE_x \EE_{y\sim \pi} [r(x,y)]$   , then from \citet{li2023policy}, we have

\begin{proposition}\label{lemma:overfit_theirs}
	$r(\pi_r^{*}) - r(\pi_{\text{PPO}}) \leq 2\eps_r + 2\eps_x$ ; $r(\pi_r^{*}) - r(\pi_{\text{DPO}}) \leq 2\eps_r + 2\eps_x + 2\eps_y $
\end{proposition}

From proposition \ref{lemma:overfit_ours}
and \ref{lemma:overfit_theirs}, we can interpret that DPO overfits on the preference dataset used for alignment and results in poor generalization, as compared to PPO. 
Next, we use results from \citet{sablayrolles2019white,aubinais2023fundamental} to lower-bound the bayes optimal membership when an algorithm is overfitting on the training dataset.

 To this end, we consider a general learning model $\pi$ that produces the parameters $\theta_n$ minimizing the loss $L_n : \theta \to \frac{1}{n} \sum_{i=1}^{n} l_{\theta}(x_i,y_i)$ for some training dataset $\{z_i:= (x_i,y_i)\}_{i=1}^n$, where $z_i$ are independent and identically sampled from the same population. Similar to \cite{sablayrolles2019white}, we assume the posterior distribution of $\theta_n$  given $\{{z_i}\}_{i=1}^{N}$ follows:
 \begin{align}
 		P({\theta}_n | z_1, \ldots, & z_n)  \propto e^{-\frac{1}{T} \sum_{i=1}^n \ell({\theta}_n, z_i)}
 	\label{eq:sdefinition}
 \end{align}
  where $T$ is a temperature parameter to factor in various randomness present in training algorithms such as sampling, dropout, etc  \citep{sablayrolles2019white,aubinais2023fundamental}. Given model parameters $\theta_n$ obtained using an algorithm $\pi$, we wish to know how much information is contained in $\theta_n$ about the memberships $m_i \in \{0,1\}$ corresponding to $z_i$ for $i \in\ \{1,\cdots,n\}$.  
Without loss of generality, let $z_1 = (x_1,y_1)$ be a prompt response pair whose membership we want to infer from an aligned model $\pi_{\theta}$. We want to determine the membership of $z_1$ i.e. $\cM (\pi_{\theta}, z_1) := P(m_1 = 1 | \pi_{\theta}, z_1)$. We collect information about the remaining data as $\ct = \{z_2,\dots,z_n\}$

We state a modified version of Theorem 2 of \cite{sablayrolles2019white} as a proposition here. For brevity, we will drop the subscript in $\theta_n$ when the context is clear.
\begin{lemma}\label{lemma:sablay_modified}
	Given the parameters $\theta$ and sample $z_1$, assuming a uniform prior over $m_1$'s, the optimal membership inference is given by: 
	\begin{align}
		\cM(\theta,z_1) = \EE_{\ct}[\sigma(s(z_1,\theta,p_{\ct}))] \label{eq:sablayrolles_modified}
	\end{align}
	where we define the corresponding posterior distribution $p_{\ct}(\theta)$, threshold $\tau_{p} $, and the score $s$ as follows: %
\begin{align}
	p_{\mathcal{T}}(\theta) &:= \frac{e^{-\frac{1}{T} \sum_{i=1}^n \ell ({\theta}, z_i)}}{\int_t e^{-\frac{1}{T} \sum_{i=1}^n  \ell (t, z_i)} \, dt} \label{eq:posterior} \\
	\tau_p(z_1) &:= T \log \left( \int_{t} e^{ \frac{1}{T}\ell(t, z_1) } p_{\mathcal{T}}(t) dt \right) 	\label{eq:taudefinition}  \\	
	s(z_1,{\theta}, p_{\mathcal{T}}) &:=  \frac{1}{T} \left( \tau_p(z_1) - \ell ({\theta}, z_1) \right) \label{eq:score}
\end{align}

%
%
\end{lemma}

The interpretation of Equation \ref{eq:sablayrolles_modified} is as follows: the lower the loss $\ell(\theta,z_1)$ becomes compared to the threshold $\tau_p(z_1)$, the more we gain non-trivial information about the membership of $z_1$. Here $\sigma$ denotes the sigmoid function. In general, the posterior $p_{\ct}$ is intractable. 

The MALT (acronym for Membership Attack Loss Threshold) assumption is a simplification framework to handle the intractability of posterior of $p_{\mathcal{T}}(\theta)$ and it assumes that the posterior $\tau_{p} := \log \left(\int_t e^{-\frac{1}{T} \ell\left(t, z_1\right)} p(t) dt \right)$ is constant \citep{sablayrolles2019white}.  Despite the simplistic assumption, it has been show that MALT has decent empirical performance \citep{,sablayrolles2019white,carlini2019secret,quan2022amplification}. Under this MALT assumption, it is possible to show that for the setting considered in Equations \ref{eq:toy_ppo_eq} and \ref{eq:toy_dpo_eq},  $M(\pi_{DPO}, z_1) \geq M(\pi_{PPO}, z_1)$. 
\begin{proposition}
	\label{prop:overfitting}
	If ${\pi}_{PPO}$ and ${\pi}_{DPO}$ are the global optimisers of the PPO and DPO objectives in Equation \ref{eq:toy_ppo_eq} and \ref{eq:toy_dpo_eq} respectively, then $M(\pi_{DPO}, z_1) \geq M(\pi_{PPO}, z_1)$ under the MALT assumption of \citet{sablayrolles2019white}. \label{prop:malt_toy_rlhf}
\end{proposition}

In general, MALT assumption is quite restrictive in nature. Hence, we don't restrict ourselves to MALT assumption and give an generalized fundamental lower bound on the score function of the optimal membership for any algorithm that overfits the training data.

Next we consider the case where the learning model $\pi$ overfits on the training data. Following \cite{aubinais2023fundamental}, we first define a probabilistic version of overfitting as follows: 

\begin{definition}
	An algorithm $\pi$ is $\mbrkt{\eps,1-\alpha}$ overfitting for some $\eps \in \mR^{+}$ and $\alpha \in (0,1)$ when $\mP\mbrkt{\ell_{\theta} (z_1) \leq \eps} \geq 1 - \alpha$.  
\end{definition}

Next, we make a simplifying assumption that under the posterior distribution given in \ref{eq:sdefinition},  the likelihoods $\{\ell_{\theta}(z_i)\}$ are independent and identically distributed.  Using this simplifying assumption, we give a sufficient condition for the above defined probabilistic overfitting based on the stopping criterion for the algorithm $\pi$.

\begin{lemma}
For some fixed $\eps \in \mR^{+}$ and $\alpha \in (0,1)$, let $\eta:= \eps \alpha$, and suppose that $\pi$ stops as soon as $L_n \leq \eta$, then $\pi$ is $\mbrkt{\eps,\mbrkt{1-\mbrkt{1-\frac{1}{n\alpha}}^{n}}}$ overfitting. \label{lemma:ours_overfitting}
\end{lemma}

Note that the probability of overfitting given in Lemma \ref{lemma:ours_overfitting} i.e. $\mbrkt{1-\mbrkt{1-\frac{1}{n\alpha}}^{n}}$ is much higher compared to the $1-\alpha$ mentioned in proposition 4.2 of \cite{aubinais2023fundamental}. Since in the case of standard MIA frameworks we are only given $z_1$ and $\theta$, we assume that $\pi$ is  $\mbrkt{\eps,\mbrkt{1-\mbrkt{1-\frac{1}{n\alpha}}^{n}}}$ overfitting given $\tau$. Under this simplifying assumption, the following theorem holds.\footnote{This assumption was added in a later revision to address a limitation in the original analysis. This modification does not affect other results or conclusions of the paper.} 

\begin{theorem}\label{thm:lowerbound}
	For some fixed $\eps$ and $\alpha \in (0,1)$, let $\eta := \eps\alpha$ and the algorithm $\pi$ stops when  $L_n(\hat{\theta}) \leq \eta$, then 
		\begin{align}
		s(z_1,{\theta}, p_{\mathcal{T}}) \geq \log\mbrkt{1-\mbrkt{1-\frac{1}{n\alpha}}^{n}} - \frac{\ell(\theta,z_1)}{T}
	\end{align}
\end{theorem}

%
%
%
%


There are two practical conclusions from Theorem \ref{thm:lowerbound} as follows: (a) keeping everything else constant, as $\alpha \to 0$ , the score function increases i.e. more the overfitting, more the susceptibility towards MIA. (b) as the number of training points $n$ increases, the score function decreases. That is, smaller the dataset size, higher the privacy risk.

From Propositions \ref{lemma:overfit_ours} and \ref{lemma:overfit_theirs}, we see that DPO is prone to overfitting, and thus, by Theorem \ref{thm:lowerbound}, it has higher susceptibility to MIA compared to PPO.  In \S \ref{sec:PREMIA}, we introduce our novel framework and  in \S \ref{sec:experiments}, we empirically show that indeed DPO has serious MIA vulnerabilities as compared to PPO.

\section{PREMIA FRAMEWORK}
\label{sec:PREMIA}
 There are two major motivations for proposing a novel optimstic attack framework: a) Traditional MIAs fail to account for the unique characteristics of preference data and the specific optimization objectives employed during the alignment of large language models (LLMs). These frameworks primarily focus on assessing attacks at the prompt-response level, neglecting the nuanced structure of preference tuples—a widely adopted data format for LLM alignment. Consequently, there is a pressing need for tailored frameworks capable of evaluating MIA not only at the prompt-response level but also at the granular preference tuple level, capturing the true extant of MIA in LLM alignment.
 
 b) Most of the attack frameworks proposed in literature are reference-free in nature, which struggle to achieve effective results when applied to LLMs which have remarkable generalization capabilities at the distributional level.  However, under the optimistic assumption of access to the base model employed for fine-tuning, it becomes feasible to design reference-based attacks, known for their superior effectiveness. Leveraging this assumption, our proposed PREMIA framework carefully crafts a reference-based attack tailored to the intricacies of  DPO and PPO — two of the widely used alignment techniques for LLMs. 
 
 The assumption of base model access, while optimistic, is not impractical in the context of LLM alignment. Many non-enterprise organizations and researchers opt to fine-tune and align open-source LLMs for their specific tasks, as closed-source models like Claude or ChatGPT often come with restrictive licenses and relatively higher inference costs. Moreover, certain government and institutional regulations prohibit the transfer of data outside specific premises, leaving local fine-tuning, alignment, and on-premise hosting of LLMs as the only viable option. By leveraging these open-source base models, practitioners can tailor the LLMs to their requirements while avoiding the limitations and expenses associated with proprietary models. Consequently, the assumption of base model access is neither entirely impractical nor over-convenient, making our proposed PREMIA framework applicable to a wide range of real-world scenarios. Further, since reference-based attacks tend to perform relatively better at detection of membership, PREMIA, with its optimistic assumption gives a practical upper bound on the extent of vulnerability.

\begin{figure}[htbp]
	\centering
\includegraphics[width=0.95\columnwidth]{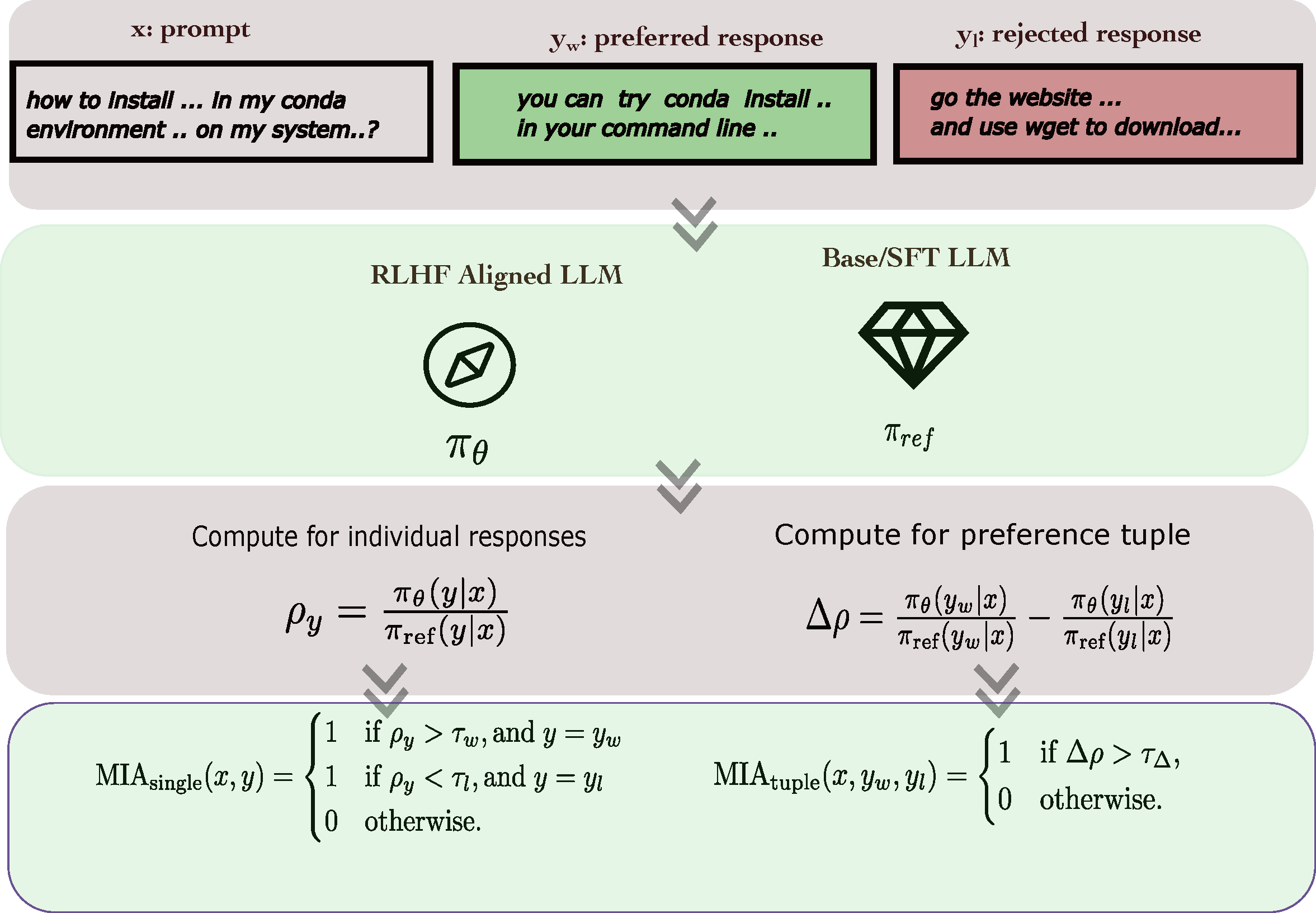}
	\label{fig:premia_overview}
	\caption{Overview of PREMIA framework for individual prompt-response pairs and the entire preference tuple}
\end{figure}

\subsection{For Individual Response}
 For individual prompt-response pairs, we define the metric as the ratio of the conditional probability of the aligned model $\pi_{\theta}$ to that of a reference model $\piref$:
\begin{equation}
	\rho_{y} = \frac{\pi_{\theta}(y|x)}{\piref(y|x)},
\end{equation}
This ratio measures the likelihood that the target model will produce a specific response compared to the reference model, indicating potential overfitting to training data. The motivation behind this specific functional form is in part due to the functional form of the reward being optimized in both PPO and DPO - refer to Equation \ref{eq:optml_ppo_reward}.

Building on top of the above metric, depending on whether the response is chosen or rejected, we define the pair's membership as:
\begin{equation}
	\text{MIA}_{\text{single}}(x, y) = 
	\begin{cases}
		1 & \text{if } \rho_{y} > \tau_w, \text{and}  \; y = y_w\\
		1 & \text{if } \rho_{y} < \tau_l, \text{and} \; y = y_l\\
		0 & \text{otherwise}.
	\end{cases}
\end{equation}
Although $\tau_y$ is mentioned, our primary metric in the experiments is the Area Under the Receiver Operating Characteristic (AUROC), which does not require setting a specific threshold. 

The choice of the reference model $\piref$ serves as a benchmark for comparing the behavior of the target model $\pi_{\theta}$. This model can be the base pre-trained model from which $\pi_{\theta}$ originated or a different base model trained on the same dataset. Our experiments, designed to test both scenarios, consistently demonstrate robust performance of our MIA method under various conditions.

\subsection{For the Entire Preference Tuple}
To ascertain the membership of the complete preference tuple $(x, y_w, y_l)$, we compute the difference between the probability ratios of the preferred and not preferred responses:
\begin{equation}
	\Delta \rho = \frac{\pi_{\theta}(y_w|x)}{\piref(y_w|x)} - \frac{\pi_{\theta}(y_l|x)}{\piref(y_l|x)}.
\end{equation}
This measure captures the comparative preference strength more effectively, offering a nuanced insight into how preference data impacts model training:
\begin{equation}
	\text{MIA}_{\text{tuple}}(x, y_w, y_l) = 
	\begin{cases}
		1 & \text{if } \Delta \rho > \tau_{\Delta},\\
		0 & \text{otherwise}.
	\end{cases}
\end{equation}

\section{EXPERIMENTS}
\label{sec:experiments}
\subsection{Research Questions}
This subsection outlines the practical research questions guiding our experimental design. More specifically we study: a) \textit{How do DPO and PPO differ in their susceptibility to Membership Inference Attacks?} b) \textit{Does model size influence its risk of data leakage through MIAs, and how does this vary between DPO and PPO trained models?} and c) \textit{What are the performance and privacy trade-offs when employing DPO versus PPO in LLMs?}.  These questions aim to evaluate the comparative effectiveness, privacy implications, and utility of DPO and PPO in aligning LLMs.

%
%

\subsection{Setup}
\label{sec:setup}
\paragraph{Models.} We conduct experiments using a variety of models to ensure a comprehensive evaluation on different scales of model complexity. We include larger models such as Gemma-2-2B \citep{gemma_2024}, Mistral-7B-v0.3, Mistral-7B-v0.1 \citep{jiang2023mistral}, Open-llama-3b and Open-llama-7b models \citep{openlm2023openllama,touvron2023llama} as well as a series of smaller models from the OpenAI GPT-2 family \citep{radford2019language}: GPT2, GPT2-medium, GPT2-large, and GPT2-xl. 
For the reference model in our ratio calculations, we primarily use the SFT model trained from the same base pre-trained version of the model being evaluated. Additionally, we conduct experiments where the reference model differs from the base model to evaluate the robustness of our methodology under varied conditions.

\paragraph{Datasets.}
For our experiments, we utilize the Stack-Exchange-Paired (SE) dataset\footnote{\url{https://huggingface.co/datasets/lvwerra/stack-exchange-paired}} and the IMDB-RLHF-Pair dataset \citep{rafailov2024direct}.
Both datasets have a prompt $x$ accompanied by two responses: the `Chosen' response $y_w$ and the `Rejected' response $y_l$.
The SE dataset contains questions and answers from the Stack Overflow dataset, where answers with more votes are preferred. The IMDB-RLHF-Pair dataset is generated by IMDB, and responses with positive sentiment are preferred.
For the Stack-Exchange-Paired dataset, the \texttt{data/rl} split is used for training, and \texttt{data/evaluation} is used as validation data. For the IMDB-RLHF-Pair dataset, 20k entries are used for training, while the remaining is for validation.

\paragraph{Evaluation Metrics.} To comprehensively assess DPO and PPO alignment, we employ a dual-focused evaluation framework encompassing utility performance and membership privacy:

\begin{itemize}
	\item \textit{Utility Performance:} Our evaluation includes the reward score of generated responses given by the reward model and perplexity for assessing fluency. We also incorporate comprehensive diversity measures: Mean Segmented Type Token Ratio (MSSTR), Distinct-1, Distinct-2, Unique-1, and Unique-2 metrics \citep{johnson1944studies, li2015diversity, ramamurthy2022reinforcement}. Additionally, we utilize advanced text generation quality metrics such as BERTScore \cite{zhang2019bertscore}, ROUGE \cite{lin2004rouge}, BLEU \cite{papineni2002bleu}, and METEOR \cite{banerjee2005meteor}, which collectively offer a nuanced view of the models' performance in terms of fluency, adequacy, and diversity, closely mirroring human judgment in text quality assessment.
	
	\item \textit{MIA Performance:} In line with previous literature on MIA \citep{shi2024detecting,zhang2024min,chen2022relaxloss}, to measure the model's susceptibility, we utilize the Area Under the Receiver Operating Characteristic curve (AUROC). This metric captures the model's defense against MIAs, reflecting the balance between true positive rate and false positive rate in identifying training data.
\end{itemize}
\paragraph{Implementation Details.} Due to the computational efficiency of LoRA, we used LoRA for all of our model training processes. Additionally, we hypothesized that fine-tuning LoRA at the RL stage would help to ensure that the aligned model does not deviate significantly from the reference model. To further improve efficiency, we also used quantization techniques. We use TRL\citep{von_werra_trl_2023} for model alignment training. More detailed implementation information can be found in Appendix~\ref{app:implementation_details}. 
\subsection{Existing MIA frameworks}
To accurately evaluate the differential susceptibility, in addition to the optimistic PREMIA, we also compare using well-known MIA frameworks such Perplexity\citep{yeom2018privacy}, Zlib\citep{carlini2021extracting}, Lowercase\citep{carlini2021extracting}, Min-k\citep{shi2023detecting}, etc. which are specifically tailored for LLMs. More details about these baselines are given in Appendix 
\ref{app:baselines_details}. These frameworks are designed to target individual prompt-response pairs of the preference data but do not extend to analyzing the entire preference tuples.

\subsubsection{Differential Susceptibility - DPO vs PPO}
\label{sec:MIA_Methodology_Effectiveness}

\begin{figure}

	\begin{minipage}{.45\textwidth}
	\centering
\captionsetup{width=0.95\linewidth}
\captionof{figure}{\small AUROC scores for $\text{MIA}_\text{Pair}$ detection for SE dataset}
\includegraphics[width=.9\linewidth]{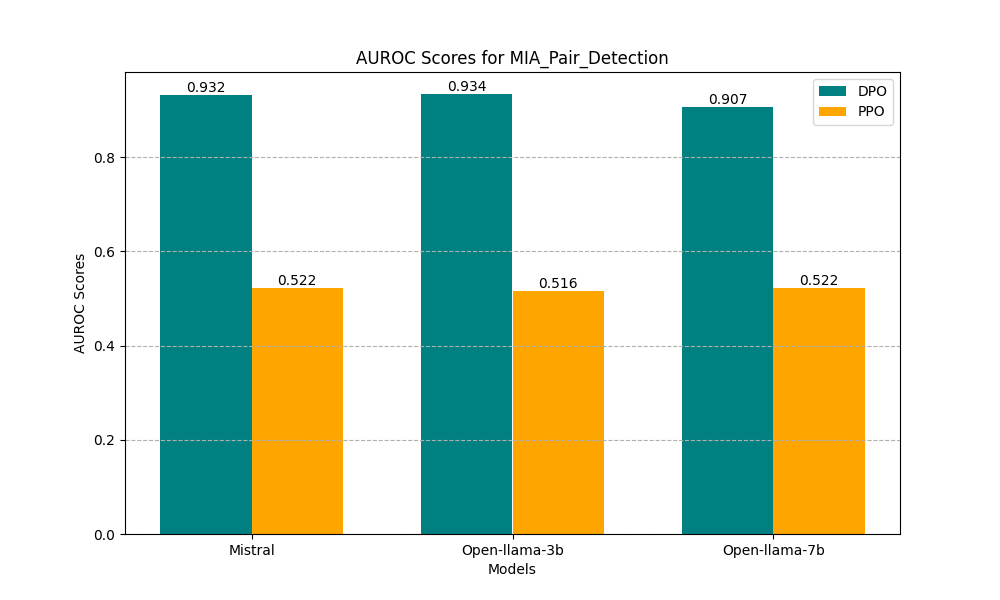}
\label{fig:MIA_Pair_BarPlot}
	\end{minipage}
	\begin{minipage}{.45\textwidth}
			\centering
\captionsetup{width=0.95\linewidth}
\captionof{figure}{Train/Eval Aaccuracy for Mistral-7B on IMDB.}

\includegraphics[width=0.8\linewidth]{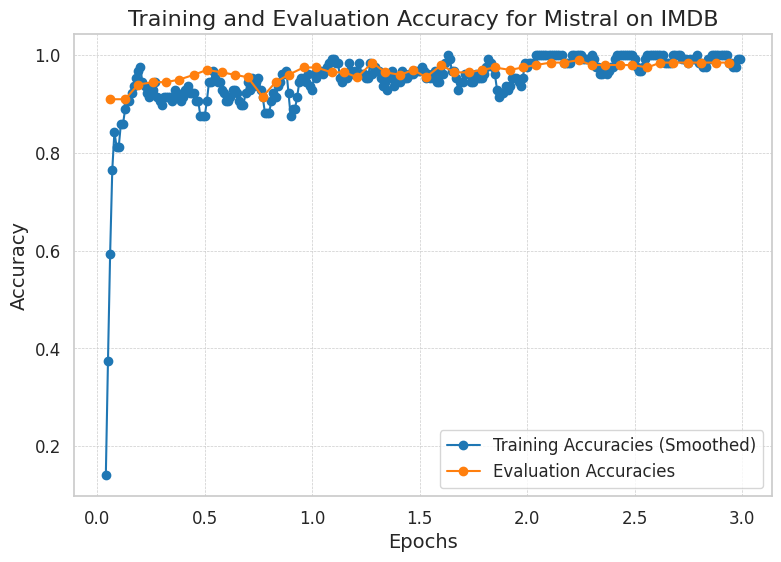}

\label{fig:acc_curve}
	\end{minipage}

\end{figure}

Table~\ref{tab:MIA_ROC_stack_imdb} presents the AUROC scores for the `Chosen' and `Rejected' responses for various MIA methods across Mistral-7B, Open-llama-3b, and Open-llama-7b models. PREMIA-base and PREMIA-SFT indicate using the base model or SFT model as the reference model respectively. In general, we find that, compared to PPO,  DPO is flagged to be highly susceptible for both the datasets across all the frameworks. 
As expected, PREMIA being an optimistic framework consistently achieves high AUROC scores.
 We do not measure the entire tuple using baselines because traditional MIA methods are not designed to handle the contextual dependencies inherent in preference tuples. AUROC numbers for PREMIA-SFT on the tuple memberships for SE dataset is given in Figure~\ref{fig:MIA_Pair_BarPlot}; Appendix \ref{app:imdb_pair_plot} has the corresponding plot for IMDB dataset.
  In order to evaluate the robustness of PREMIA, we switch the reference models within the same family and evaluate the AUROC. As shown in Table \ref{tab:switch_ref_model}, this did not deteriorate the performance significantly.

\definecolor{JapaneseLaurel}{rgb}{0,0.501,0}
\definecolor{WebOrange}{rgb}{1,0.647,0}
\captionsetup{font=small}
\begin{table}[htp]
	\tiny
	\centering
	\captionof{table}{\small AUROC scores comparing different MIA methods on Gemma-2-2B, Mistral-7B-v\{0.1,0.3\}, Open-llama-3b, and Open-llama-7b models are presented, where higher scores indicate greater susceptibility to MIA. \textcolor{WebOrange}{The best} and \textcolor{JapaneseLaurel}{second-best} scores in each column are highlighted in orange and green, respectively. \uline{The better} score between DPO and PPO trained models is underlined.}
	\label{tab:MIA_ROC_stack_imdb}
\begin{tblr}{
  row{4} = {c},
  row{5} = {c},
  row{6} = {c},
  row{7} = {c},
  row{8} = {c},
  row{9} = {c},
  row{10} = {c},
  row{11} = {c},
  row{12} = {c},
  row{13} = {c},
  row{14} = {c},
  row{15} = {c},
  row{16} = {c},
  row{17} = {c},
  row{18} = {c},
  row{19} = {c},
  row{20} = {c},
  row{21} = {c},
  row{22} = {c},
  row{23} = {c},
  row{24} = {c},
  cell{1}{3} = {c=4}{c},
  cell{1}{7} = {c=4}{c},
  cell{2}{2} = {c},
  cell{2}{3} = {c=2}{c},
  cell{2}{5} = {c=2}{c},
  cell{2}{7} = {c=2}{c},
  cell{2}{9} = {c=2}{c},
  cell{3}{2} = {c},
  cell{3}{3} = {c},
  cell{3}{4} = {c},
  cell{3}{5} = {c},
  cell{3}{6} = {c},
  cell{3}{7} = {c},
  cell{3}{8} = {c},
  cell{3}{9} = {c},
  cell{3}{10} = {c},
  cell{4}{1} = {r=8}{},
  cell{4}{2} = {r},
  cell{5}{2} = {r},
  cell{5}{3} = {fg=JapaneseLaurel},
  cell{5}{5} = {fg=JapaneseLaurel},
  cell{6}{2} = {r},
  cell{6}{10} = {fg=JapaneseLaurel},
  cell{7}{2} = {r},
  cell{8}{2} = {r},
  cell{8}{8} = {fg=JapaneseLaurel},
  cell{8}{10} = {fg=WebOrange},
  cell{9}{2} = {r},
  cell{10}{2} = {r},
  cell{10}{4} = {fg=JapaneseLaurel},
  cell{10}{6} = {fg=WebOrange},
  cell{10}{7} = {fg=JapaneseLaurel},
  cell{10}{9} = {fg=JapaneseLaurel},
  cell{11}{2} = {r},
  cell{11}{3} = {fg=WebOrange},
  cell{11}{4} = {fg=WebOrange},
  cell{11}{5} = {fg=WebOrange},
  cell{11}{6} = {fg=JapaneseLaurel},
  cell{11}{7} = {fg=WebOrange},
  cell{11}{8} = {fg=WebOrange},
  cell{11}{9} = {fg=WebOrange},
  cell{11}{10} = {fg=JapaneseLaurel},
  cell{12}{1} = {r=8}{},
  cell{12}{2} = {r},
  cell{13}{2} = {r},
  cell{13}{3} = {fg=JapaneseLaurel},
  cell{13}{5} = {fg=JapaneseLaurel},
  cell{14}{2} = {r},
  cell{15}{2} = {r},
  cell{16}{2} = {r},
  cell{17}{2} = {r},
  cell{17}{9} = {fg=WebOrange},
  cell{18}{2} = {r},
  cell{18}{4} = {fg=JapaneseLaurel},
  cell{18}{6} = {fg=JapaneseLaurel},
  cell{18}{7} = {fg=JapaneseLaurel},
  cell{18}{8} = {fg=JapaneseLaurel},
  cell{18}{10} = {fg=JapaneseLaurel},
  cell{19}{2} = {r},
  cell{19}{3} = {fg=WebOrange},
  cell{19}{4} = {fg=WebOrange},
  cell{19}{5} = {fg=WebOrange},
  cell{19}{6} = {fg=WebOrange},
  cell{19}{7} = {fg=WebOrange},
  cell{19}{8} = {fg=WebOrange},
  cell{19}{9} = {fg=JapaneseLaurel},
  cell{19}{10} = {fg=WebOrange},
  cell{20}{1} = {r=8}{},
  cell{20}{2} = {r},
  cell{20}{4} = {fg=JapaneseLaurel},
  cell{21}{2} = {r},
  cell{21}{3} = {fg=WebOrange},
  cell{21}{4} = {fg=WebOrange},
  cell{21}{5} = {fg=JapaneseLaurel},
  cell{21}{6} = {fg=WebOrange},
  cell{21}{8} = {fg=WebOrange},
  cell{22}{2} = {r},
  cell{22}{8} = {fg=JapaneseLaurel},
  cell{22}{10} = {fg=JapaneseLaurel},
  cell{23}{2} = {r},
  cell{23}{5} = {fg=WebOrange},
  cell{24-43}{2} = {r},
  cell{25}{3} = {c},
  cell{25}{4} = {c},
  cell{25}{5} = {c},
  cell{25}{6} = {c},
  cell{25}{7} = {c},
  cell{25}{8} = {c},
  cell{25}{9} = {c},
  cell{25}{10} = {c},
  cell{26}{3} = {c},
  cell{26}{4} = {c},
  cell{26}{5} = {c},
  cell{26}{6} = {c},
  cell{26}{7} = {c,fg=JapaneseLaurel},
  cell{26}{8} = {c,fg=JapaneseLaurel},
  cell{26}{9} = {c,fg=WebOrange},
  cell{26}{10} = {c,fg=WebOrange},
  cell{27}{3} = {c,fg=JapaneseLaurel},
  cell{27}{4} = {c},
  cell{27}{5} = {c},
  cell{27}{6} = {c,fg=JapaneseLaurel},
  cell{27}{7} = {c,fg=WebOrange},
  cell{27}{8} = {c},
  cell{27}{9} = {c,fg=JapaneseLaurel},
  cell{27}{10} = {c,fg=JapaneseLaurel},
  cell{28}{1} = {r=8}{},
  cell{28}{3} = {c},
  cell{28}{4} = {c},
  cell{28}{5} = {c},
  cell{28}{6} = {c},
  cell{28}{7} = {c},
  cell{28}{8} = {c},
  cell{28}{9} = {c},
  cell{28}{10} = {c},
  cell{29}{3} = {c,fg=WebOrange},
  cell{29}{4} = {c,fg=JapaneseLaurel},
  cell{29}{5} = {c,fg=JapaneseLaurel},
  cell{29}{6} = {c},
  cell{29}{7} = {c},
  cell{29}{8} = {c},
  cell{29}{9} = {c},
  cell{29}{10} = {c},
  cell{30}{3} = {c},
  cell{30}{4} = {c,fg=WebOrange},
  cell{30}{5} = {c},
  cell{30}{6} = {c,fg=WebOrange},
  cell{30}{7} = {c},
  cell{30}{8} = {c},
  cell{30}{9} = {c},
  cell{30}{10} = {c},
  cell{31}{3} = {c},
  cell{31}{4} = {c},
  cell{31}{5} = {c},
  cell{31}{6} = {c},
  cell{31}{7} = {c},
  cell{31}{8} = {c},
  cell{31}{9} = {c},
  cell{31}{10} = {c},
  cell{32}{3} = {c},
  cell{32}{4} = {c},
  cell{32}{5} = {c},
  cell{32}{6} = {c},
  cell{32}{7} = {c},
  cell{32}{8} = {c},
  cell{32}{9} = {c},
  cell{32}{10} = {c},
  cell{33}{3} = {c},
  cell{33}{4} = {c},
  cell{33}{5} = {c},
  cell{33}{6} = {c},
  cell{33}{7} = {c},
  cell{33}{8} = {c},
  cell{33}{9} = {c},
  cell{33}{10} = {c},
  cell{34}{3} = {c},
  cell{34}{4} = {c},
  cell{34}{5} = {c},
  cell{34}{6} = {c,fg=JapaneseLaurel},
  cell{34}{7} = {c,fg=JapaneseLaurel},
  cell{34}{8} = {c,fg=WebOrange},
  cell{34}{9} = {c,fg=JapaneseLaurel},
  cell{34}{10} = {c,fg=JapaneseLaurel},
  cell{35}{3} = {c,fg=JapaneseLaurel},
  cell{35}{4} = {c},
  cell{35}{5} = {c,fg=WebOrange},
  cell{35}{6} = {c},
  cell{35}{7} = {c,fg=WebOrange},
  cell{35}{8} = {c,fg=JapaneseLaurel},
  cell{35}{9} = {c,fg=WebOrange},
  cell{35}{10} = {c,fg=WebOrange},
  cell{36}{1} = {r=8}{},
  cell{36}{3} = {c},
  cell{36}{4} = {c,fg=JapaneseLaurel},
  cell{36}{5} = {c},
  cell{36}{6} = {c},
  cell{36}{7} = {c},
  cell{36}{8} = {c},
  cell{36}{9} = {c},
  cell{36}{10} = {c,fg=JapaneseLaurel},
  cell{37}{3} = {c,fg=WebOrange},
  cell{37}{4} = {c,fg=WebOrange},
  cell{37}{5} = {c,fg=JapaneseLaurel},
  cell{37}{6} = {c,fg=JapaneseLaurel},
  cell{37}{7} = {c},
  cell{37}{8} = {c},
  cell{37}{9} = {c},
  cell{37}{10} = {c},
  cell{38}{3} = {c},
  cell{38}{4} = {c},
  cell{38}{5} = {c},
  cell{38}{6} = {c},
  cell{38}{7} = {c},
  cell{38}{8} = {c},
  cell{38}{9} = {c},
  cell{38}{10} = {c},
  cell{39}{3} = {c},
  cell{39}{4} = {c},
  cell{39}{5} = {c},
  cell{39}{6} = {c},
  cell{39}{7} = {c},
  cell{39}{8} = {c},
  cell{39}{9} = {c},
  cell{39}{10} = {c},
  cell{40}{3} = {c,fg=JapaneseLaurel},
  cell{40}{4} = {c},
  cell{40}{5} = {c},
  cell{40}{6} = {c},
  cell{40}{7} = {c},
  cell{40}{8} = {c},
  cell{40}{9} = {c},
  cell{40}{10} = {c,fg=WebOrange},
  cell{41}{3} = {c},
  cell{41}{4} = {c},
  cell{41}{5} = {c},
  cell{41}{6} = {c},
  cell{41}{7} = {c},
  cell{41}{8} = {c,fg=WebOrange},
  cell{41}{9} = {c},
  cell{41}{10} = {c},
  cell{42}{3} = {c},
  cell{42}{4} = {c},
  cell{42}{5} = {c},
  cell{42}{6} = {c},
  cell{42}{7} = {c,fg=JapaneseLaurel},
  cell{42}{8} = {c,fg=JapaneseLaurel},
  cell{42}{9} = {c,fg=WebOrange},
  cell{42}{10} = {c},
  cell{43}{3} = {c},
  cell{43}{4} = {c},
  cell{43}{5} = {c,fg=WebOrange},
  cell{43}{6} = {c,fg=WebOrange},
  cell{43}{7} = {c,fg=WebOrange},
  cell{43}{8} = {c},
  cell{43}{9} = {c,fg=JapaneseLaurel},
  cell{43}{10} = {c},
  vline{3-4} = {1}{},
  vline{3-4,6,8} = {2}{},
  vline{3,5,7,9} = {3-43}{},
  hline{1,4} = {-}{},
  hline{2-3} = {3-10}{},
  hline{12,20,28,36,44} = {2-10}{},
  colsep=2.3pt,
}
                                              &             & IMDB                       &               &                              &               & Stack-Exchange             &       &                              &       \\
                                              &             & $\text{MIA}_\text{Chosen}$ &               & $\text{MIA}_\text{Rejected}$ &               & $\text{MIA}_\text{Chosen}$ &       & $\text{MIA}_\text{Rejected}$ &       \\
                                              &             & DPO                        & PPO           & DPO                          & PPO           & DPO                        & PPO   & DPO                          & PPO   \\
\begin{sideways}Gemma-2-2B\end{sideways}      & PPL         & \uline{0.572}              & 0.531         & \uline{0.575}                & 0.541         & \uline{0.557}              & 0.505 & \uline{0.542}                & 0.517 \\
                                              & Zlib        & \uline{0.589}              & 0.532         & \uline{0.591}                & 0.531         & \uline{0.579}              & 0.517 & \uline{0.589}                & 0.534 \\
                                              & Lowercase   & \uline{0.556}              & 0.538         & \uline{0.545}                & 0.525         & \uline{0.565}              & 0.531 & \uline{0.556}                & 0.547 \\
                                              & Ref         & \uline{0.570}              & 0.530         & \uline{0.576}                & 0.547         & \uline{0.552}              & 0.527 & \uline{0.570}                & 0.536 \\
                                              & MIN-K       & \uline{0.571}              & 0.524         & \uline{0.587}                & 0.523         & \uline{0.605}              & 0.538 & \uline{0.571}                & 0.551 \\
                                              & N-hood      & \uline{0.575}              & 0.512         & \uline{0.568}                & 0.513         & \uline{0.582}              & 0.519 & \uline{0.632}                & 0.521 \\
                                              & PREMIA-base & \uline{0.581}              & 0.539         & \uline{0.585}                & 0.555         & \uline{0.691}              & 0.534 & \uline{0.671}                & 0.545 \\
                                              & PREMIA-SFT  & \uline{0.592}              & 0.545         & \uline{0.597}                & 0.552         & \uline{0.714}              & 0.539 & \uline{0.681}                & 0.547 \\
\begin{sideways}Mistral-7B-v0.3\end{sideways} & PPL         & \uline{0.592}              & 0.525         & \uline{0.592}                & 0.525         & \uline{0.575}              & 0.521 & \uline{0.592}                & 0.525 \\
                                              & Zlib        & \uline{0.598}              & 0.521         & \uline{0.598}                & 0.521         & \uline{0.584}              & 0.529 & \uline{0.598}                & 0.521 \\
                                              & Lowercase   & \uline{0.567}              & 0.505         & \uline{0.567}                & 0.505         & \uline{0.581}              & 0.514 & \uline{0.567}                & 0.505 \\
                                              & Ref         & \uline{0.587}              & 0.538         & \uline{0.587}                & 0.538         & \uline{0.572}              & 0.523 & \uline{0.587}                & 0.538 \\
                                              & MIN-K       & \uline{0.589}              & 0.534         & \uline{0.589}                & 0.534         & \uline{0.578}              & 0.508 & \uline{0.589}                & 0.534 \\
                                              & N-hood      & \uline{0.576}              & 0.509         & \uline{0.577}                & 0.512         & \uline{0.582}              & 0.519 & \uline{0.632}                & 0.514 \\
                                              & PREMIA-base & \uline{0.596}              & 0.543         & \uline{0.596}                & 0.543         & \uline{0.758}              & 0.531 & \uline{0.596}                & 0.543 \\
                                              & PREMIA-SFT  & \uline{0.619}              & 0.558         & \uline{0.619}                & 0.558         & \uline{0.789}              & 0.543 & \uline{0.619}                & 0.558 \\
\begin{sideways}Mistral-7B-v0.1\end{sideways}      & PPL         & \uline{0.569}              & 0.538         & \uline{0.588}                & 0.503         & \uline{0.575}              & 0.512 & \uline{0.586}                & 0.515 \\
                                              & Zlib        & \uline{0.593}              & 0.568         & \uline{0.606}                & 0.536         & \uline{0.539}              & 0.529 & \uline{0.553}                & 0.516 \\
                                              & Lowercase   & \uline{0.516}              & 0.509         & \uline{0.515}                & 0.501         & \uline{0.571}              & 0.523 & \uline{0.574}                & 0.520 \\
                                              & Ref         & \uline{0.571}              & 0.533         & \uline{0.607}                & 0.511         & \uline{0.572}              & 0.518 & \uline{0.589}                & 0.512 \\
                                              & MIN-K       & \uline{0.564}              & 0.511         & \uline{0.582}                & 0.509         & \uline{0.582}              & 0.519 & \uline{0.632}                & 0.514 \\
                                              & N-hood      & \uline{0.548}              & 0.504         & \uline{0.528}                & 0.516         & \uline{0.600}              & 0.513 & \uline{0.568}                & 0.506 \\
                                              & PREMIA-base & \uline{0.571}              & 0.523         & \uline{0.592}                & 0.517         & \uline{0.771}              & 0.523 & \uline{0.764}                & 0.526 \\
                                              & PREMIA-SFT  & \uline{0.576}              & 0.507         & \uline{0.590}                & 0.527         & \uline{0.803}              & 0.521 & \uline{0.760}                & 0.520 \\
\begin{sideways}Open-llama-3b\end{sideways}   & PPL         & \uline{0.580}              & 0.508         & \uline{0.573}                & 0.526         & \uline{0.574}              & 0.515 & \uline{0.544}                & 0.512 \\
                                              & Zlib        & \uline{0.602}              & 0.540         & \uline{0.595}                & 0.506         & \uline{0.539}              & 0.517 & \uline{0.514}                & 0.521 \\
                                              & Lowercase   & 0.541                      & \uline{0.556} & 0.546                        & \uline{0.551} & \uline{0.633}              & 0.517 & \uline{0.577}                & 0.523 \\
                                              & Ref         & \uline{0.587}              & 0.508         & \uline{0.590}                & 0.535         & \uline{0.596}              & 0.524 & \uline{0.582}                & 0.518 \\
                                              & MIN-K       & \uline{0.587}              & 0.523         & \uline{0.579}                & 0.529         & \uline{0.578}              & 0.524 & \uline{0.599}                & 0.516 \\
                                              & N-hood      & \uline{0.567}              & 0.511         & \uline{0.586}                & 0.521         & \uline{0.601}              & 0.523 & \uline{0.583}                & 0.520 \\
                                              & PREMIA-base & \uline{0.564}              & 0.520         & \uline{0.562}                & 0.540         & \uline{0.758}              & 0.531 & \uline{0.750}                & 0.525 \\
                                              & PREMIA-SFT  & \uline{0.594}              & 0.504         & \uline{0.609}                & 0.518         & \uline{0.789}              & 0.525 & \uline{0.761}                & 0.534 \\
\begin{sideways}Open-llama-7b\end{sideways}   & PPL         & \uline{0.577}              & 0.529         & \uline{0.572}                & 0.505         & \uline{0.577}              & 0.514 & \uline{0.551}                & 0.527 \\
                                              & Zlib        & \uline{0.599}              & 0.559         & \uline{0.593}                & 0.525         & \uline{0.535}              & 0.509 & \uline{0.531}                & 0.516 \\
                                              & Lowercase   & \uline{0.537}              & 0.501         & \uline{0.540}                & 0.502         & \uline{0.548}              & 0.516 & \uline{0.569}                & 0.518 \\
                                              & Ref         & \uline{0.583}              & 0.515         & \uline{0.586}                & 0.503         & \uline{0.605}              & 0.516 & \uline{0.580}                & 0.517 \\
                                              & MIN-K       & \uline{0.597}              & 0.527         & \uline{0.583}                & 0.511         & \uline{0.567}              & 0.510 & \uline{0.590}                & 0.530 \\
                                              & N-hood      & \uline{0.563}              & 0.507         & \uline{0.576}                & 0.517         & \uline{0.598}              & 0.525 & \uline{0.576}                & 0.514 \\
                                              & PREMIA-base & \uline{0.559}              & 0.511         & \uline{0.560}                & 0.504         & \uline{0.761}              & 0.523 & \uline{0.759}                & 0.516 \\
                                              & PREMIA-SFT  & \uline{0.594}              & 0.511         & \uline{0.611}                & 0.527         & \uline{0.774}              & 0.516 & \uline{0.730}                & 0.524 
\end{tblr}
\end{table}

\begin{table}[h]
	\tiny
	\centering
	\captionsetup{width=0.95\linewidth}
	\captionof{table}{\small Privacy vs Utility Trade-off analysis on the Mistral-7B model on SE dataset. \label{tab:trade_off}}
	\begin{tblr}{
			row{even} = {c},
			row{3} = {c},
			row{5} = {c},
			row{7} = {c},
			row{9} = {c},
			row{11} = {c},
			row{13} = {c},
			row{15} = {c},
			cell{1}{2} = {c},
			cell{1}{3} = {c},
			cell{1}{4} = {c},
			cell{1}{5} = {c},
			cell{2}{1} = {r},
			cell{3}{1} = {r},
			cell{4}{1} = {r},
			cell{5}{1} = {r},
			cell{6}{1} = {r},
			cell{7}{1} = {r},
			cell{8}{1} = {r},
			cell{9}{1} = {r},
			cell{10}{1} = {r},
			cell{11}{1} = {r},
			cell{12}{1} = {r},
			cell{13}{1} = {r},
			cell{14}{1} = {r},
			cell{15}{1} = {r},
			vline{2-5} = {-}{},
			hline{1,16} = {-}{0.08em},
			hline{2} = {-}{0.05em},
			hline{5-6} = {-}{},
		}
		& Base   & SFT    & PPO    & DPO    \\
		$\text{MIA}_{\text{Chosen}}$   & —      & 0.53   & 0.52   & 0.80   \\
		$\text{MIA}_{\text{Rejected}}$ & —      & 0.61   & 0.52   & 0.76   \\
		$\text{MIA}_{\text{Pair}}$     & —      & 0.55   & 0.52  & 0.93   \\
		Reward$\uparrow$       & -1.922 & -1.953 & -0.771 & -1.035 \\
		PPL$\downarrow$           & 11.148 & 7.673  & 11.671 & 14.991 \\
		msttr-100$\uparrow$     & 0.673  & 0.651  & 0.633  & 0.640  \\
		distinct 1$\uparrow$    & 0.180  & 0.127  & 0.085  & 0.123  \\
		distinct 2$\uparrow$    & 0.631  & 0.521  & 0.422  & 0.520  \\
		unique 1$\uparrow$      & 2010   & 3213   & 3530   & 3059   \\
		unique 2$\uparrow$      & 9507   & 17238  & 25205  & 18017  \\
		Bert Score$\uparrow$    & 0.876  & 0.879  & 0.883  & 0.877  \\
		ROUGE$\uparrow$         & 0.424  & 0.458  & 0.457  & 0.443  \\
		BLEU$\uparrow$          & 0.348  & 0.367  & 0.338  & 0.360  \\
		METEOR$\uparrow$        & 0.445  & ~0.467 & 0.449  & 0.466  
	\end{tblr}
\end{table}

\begin{table}[h]
		\small
	\centering
	\captionof{table}{\small PREMIA of Openllama-7b when Openllama-3b model is used as reference model.}
	\label{tab:switch_ref_model}
	
	\begin{tabular}{c|c|c|c|c}
		\hline
		dataset &   $\pi_{\theta}/ \pi_{ref}$               & chosen & rejected & pair \\ \hline \hline
		\multirow{2}{*}{SE}   & open-llama-3b/7b &  0.754      &  0.673        & 0.863     \\
		& open-llama-7b/3b &  0.722      &  0.732        & 0.882     \\ \hline
		\multirow{2}{*}{IMDB} & open-llama-3b/7b &   0.585     &  0.592        &  0.512    \\
		& open-llama-7b/3b &  0.581      &  0.590        &  0.508    \\
	\end{tabular}
\end{table}

\subsubsection{Impact of Model Size on MIA Effectiveness}
\label{sec:effect_of_size}

 Table~\ref{tab:MIA_gpt2} details the PREMIA-SFT results for models of different sizes in the GPT series on the SE and IMDB datasets; refer Appendix \ref{app:GPT2_exp} for results from other frameworks.  \textcolor{black}{First, we note that in the GPT2 models as well DPO demonstrates a higher vulnerability compared to PPO. Looking at the AUROC values in both Table \ref{tab:MIA_ROC_stack_imdb} and \ref{tab:MIA_gpt2} we note that PPO aligned models are nearly impregnable to MIA with all of the existing frameworks. Second, we observe that, when compared to Mistral-7B and OpenLlama models, GPT2 models aligned using DPO show higher vulnerability on both the tasks. When compared between the tasks, we note that vulnerability is higher for SE dataset compared to IMDB dataset. This is because IMDB task is relatively easier (wherein the model has has to generate positive responses), compared to SE task (the model has to produce an answer for an SE question). As shown in Fig.~\ref{fig:acc_curve}, Mistral-7B aligned with DPO achieves over 90\% accuracy in distinguishing between selected and rejected responses in only the first 0.2 epoch for the IMDB dataset. Large pre-trained models like Mistral-7B already have strong generalization capabilities, which undermines the effectiveness of MIA. Similarly, large GPT2 models such as GPT2-xl show better generalization on simple tasks, making them less susceptible to MIA \citep{tanzer2021memorisation,wei2024memorization}. Note that this highlights a new understanding on how MIA varies with model size - on pre-training data, \citet{shi2024detecting} found that model vulnerability increases with model size for MIA; whereas for preference data, we notice that the relationship is a bit more nuanced, depending on the task complexity during alignment.} 

\subsubsection{Trade-Off between Performance and Privacy}
\label{sec:Performance_Privacy_Trade_Off}
Table~\ref{tab:trade_off} analyzes the trade-off between vulnerability to MIA and model utility between DPO and PPO for Mistral-7B on SE dataset. The "Reward" row represents the average reward score given by the reward model for each of these models, indicating how well the task was accomplished. Clearly, DPO and PPO have better reward and utility metrics compared to the rest. Further, DPO is clearly more vulnerable to MIA. 
It is worth noting that PPO provides similar utility performance to DPO, but it has a lower AUROC. These findings are in line with existing research, which also shows that despite DPO being relatively straightforward to train, it does not improve the model performance compared to PPO \cite{ivison2024unpacking,xu2024dpo}. Anecdotal responses from both these datasets to various alignment approaches are given in Appendix \ref{app:anecdotal_responses}.

\begin{table}[H]
	\tiny
	\centering
	\captionof{table}{\small Performance of PREMIA-SFT on various GPT2 model variants across SE and IMDB datasets.}
	\label{tab:MIA_gpt2}
	\begin{tblr}{
			row{4} = {c},
			row{5} = {c},
			row{6} = {c},
			row{7} = {c},
			row{8} = {c},
			row{9} = {c},
			row{11} = {c},
			row{12} = {c},
			row{13} = {c},
			row{14} = {c},
			row{15} = {c},
			row{16} = {c},
			cell{1}{2} = {c},
			cell{1}{3} = {c=2}{c},
			cell{1}{5} = {c=2}{c},
			cell{1}{7} = {c=2}{c},
			cell{2}{3} = {c},
			cell{2}{4} = {c},
			cell{2}{5} = {c},
			cell{2}{6} = {c},
			cell{2}{7} = {c},
			cell{2}{8} = {c},
			cell{3}{1} = {r=7}{},
			cell{3}{2} = {r},
			cell{3}{3} = {c},
			cell{3}{4} = {c},
			cell{3}{5} = {c},
			cell{3}{6} = {c},
			cell{3}{7} = {c},
			cell{3}{8} = {c},
			cell{4}{2} = {r},
			cell{5}{2} = {r},
			cell{6}{2} = {r},
			cell{7}{2} = {r},
			cell{8}{2} = {r},
			cell{9}{2} = {r},
			cell{10}{1} = {r=7}{},
			cell{10}{2} = {r},
			cell{10}{3} = {c},
			cell{10}{4} = {c},
			cell{10}{5} = {c},
			cell{10}{6} = {c},
			cell{10}{7} = {c},
			cell{10}{8} = {c},
			cell{11}{2} = {r},
			cell{12}{2} = {r},
			cell{13}{2} = {r},
			cell{14}{2} = {r},
			cell{15}{2} = {r},
			cell{16}{2} = {r},
			vline{3-4,6} = {1}{},
			vline{3,5,7} = {2-16}{},
			hline{1-2,10} = {-}{},
			hline{3,7,9,14,16-17} = {2-8}{},
			hline{4-6,8,11-13,15} = {2-8}{dashed},
			colsep=2.3pt,
		}
		&               & $\text{MIA}_\text{Chosen}$ &                 & $\text{MIA}_\text{Rejected}$ &                 & $\text{MIA}_\text{Pair}$ &                 \\
		&               & DPO                        & PPO             & DPO                          & PPO             & DPO                      & PPO             \\
		\begin{sideways}Stack Exchange\end{sideways} & GPT2          & 0.824            & 0.513 & 0.749              & 0.510 & 0.899          & 0.515 \\
		& GPT2-medium   & 0.829            & 0.518 & 0.752              & 0.530 & 0.910          & 0.536 \\
		& GPT2-large    & 0.857            & 0.508 & 0.758              & 0.519 & 0.916          & 0.517 \\
		& GPT2-xl       & 0.858            & 0.516 & 0.788              & 0.509 & 0.928          & 0.519 \\
		& Open-llama-3b & ~0.789           & 0.525 & 0.761              & 0.534 & 0.934          & 0.516 \\
		& Open-llama-7b & 0.774            & 0.515 & 0.730              & 0.523 & 0.907          & 0.526 \\
		& Mistral-7B    & 0.803            & 0.520 & 0.760              & 0.520 & 0.932          & 0.522 \\
		\begin{sideways}IMDB\end{sideways}           & GPT2          & 0.636                      & 0.550           & 0.713                        & 0.511           & 0.771                    & 0.549           \\
		& GPT2-medium   & 0.641                      & 0.549           & 0.707                        & 0.539           & 0.762                    & 0.528           \\
		& GPT2-large    & 0.615                      & 0.611           & 0.659                        & 0.583           & 0.704                    & 0.520           \\
		& GPT2-xl       & 0.623                      & 0.591           & 0.643                        & 0.579           & 0.692                    & 0.519           \\
		& Open-llama-3b & 0.594                      & 0.503 & 0.609                        & 0.518 & 0.509                    & 0.517 \\
		& Open-llama-7b & 0.594                      & 0.510 & 0.611                        & 0.526 & 0.500                    & 0.512 \\
		& Mistral-7B    & 0.572                      & 0.507 & 0.611                        & 0.527 & 0.556                    & 0.513 
	\end{tblr}
\end{table}


%

\section{CONCLUSION AND LIMITATIONS}

This paper examines the vulnerability of preference datasets in LLM alignment to MIAs. We show both theoretically and empirically that models trained with DPO are more susceptible to MIAs than those using PPO. Further, we also discover that the extent of MIA depends on various factors such as model size, task complexity, etc. While we have focused on highlighting MIA vulnerabilites, we haven't touched upon the various ways to mitigate them. Techniques such as DP-SGD \cite{abadi2016deep}, model pruning \citep{han2015deep}, and privacy-aware losses \citep{chen2022relaxloss} could be evaluated in the context of MIA on preference data. While optimistic frameworks such as PREMIA work for opensource models, there is a need for designing effective frameworks for closed-source LLMs where there is no access to the base model. 

\bibliography{iclr2024_conference}
\bibliographystyle{apalike}

\section*{Checklist}
\begin{enumerate}

	\item For all models and algorithms presented, check if you include:
	\begin{enumerate}
		\item A clear description of the mathematical setting, assumptions, algorithm, and/or model. [Yes]
		\item An analysis of the properties and complexity (time, space, sample size) of any algorithm. [Not Applicable]
		\item (Optional) Anonymized source code, with specification of all dependencies, including external libraries. [Yes] - as footnotes in \S \ref{sec:setup}
	\end{enumerate}

	\item For any theoretical claim, check if you include:
	\begin{enumerate}
		\item Statements of the full set of assumptions of all theoretical results. [Yes]
		\item Complete proofs of all theoretical results. [Yes] - in Appendix \ref{app:proofs}
		\item Clear explanations of any assumptions. [Yes]     
	\end{enumerate}

	\item For all figures and tables that present empirical results, check if you include:
	\begin{enumerate}
		\item The code, data, and instructions needed to reproduce the main experimental results (either in the supplemental material or as a URL). [Yes]
		\item All the training details (e.g., data splits, hyperparameters, how they were chosen). [Yes]
		\item A clear definition of the specific measure or statistics and error bars (e.g., with respect to the random seed after running experiments multiple times). [Yes]
		\item A description of the computing infrastructure used. (e.g., type of GPUs, internal cluster, or cloud provider). [Yes]
	\end{enumerate}
	
	\item If you are using existing assets (e.g., code, data, models) or curating/releasing new assets, check if you include:
	\begin{enumerate}
		\item Citations of the creator If your work uses existing assets. [Yes]
		\item The license information of the assets, if applicable. [Not Applicable]
		\item New assets either in the supplemental material or as a URL, if applicable. [Not Applicable]
		\item Information about consent from data providers/curators. [Not Applicable]
		\item Discussion of sensible content if applicable, e.g., personally identifiable information or offensive content. [Not Applicable]
	\end{enumerate}
	
	\item If you used crowdsourcing or conducted research with human subjects, check if you include:
	\begin{enumerate}
		\item The full text of instructions given to participants and screenshots. [Not Applicable]
		\item Descriptions of potential participant risks, with links to Institutional Review Board (IRB) approvals if applicable. [Not Applicable]
		\item The estimated hourly wage paid to participants and the total amount spent on participant compensation. [Not Applicable]
	\end{enumerate}
	
\end{enumerate}

\appendix
\newpage
%
%
%





%

%

\onecolumn
\aistatstitle{Instructions for Paper Submissions to AISTATS 2025: \\
Supplementary Materials}	

\addcontentsline{toc}{section}{Appendix} 
\part{Appendix} 
\parttoc 
\section{DETAILS OF BASELINES}
\label{app:baselines_details}
\paragraph{Perplexity (PPL):}
The loss attack method, based on the approach outlined in \cite{yeom2018privacy}, utilizes the perplexity of a sequence to gauge how well a language model predicts the tokens within that sequence. Perplexity is defined as:
\begin{equation}
	\mathcal{P} = \exp\left(-\frac{1}{n}\sum_{i = 1}^n \log \pi_\theta(x_i | x_1, \ldots, x_{i - 1})\right),
\end{equation}
where a lower perplexity indicates a higher likelihood that the sequence was the training data.

\paragraph{Comparing to zlib Compression (Zlib):}
This method measures the entropy of a sequence when compressed using zlib, compares the perplexity of a model to its zlib compression entropy, and uses their ratio as an inference metric ~\cite{carlini2021extracting}.

\paragraph{Comparing to Lowercased Text (Lowercase):}
This method evaluates the change in perplexity of a sequence before and after it has been lowercased, to assess the model's dependency on specific capitalization~\cite{carlini2021extracting}:
\begin{equation}
	\text{Perplexity Ratio} = \frac{\mathcal{P}(\text{Original})}{\mathcal{P}(\text{Lowercased})}.
\end{equation}

\paragraph{Comparing to Other Neural Language Models (Ref):}
This approach consists of comparing the ease of error of sequences between the target model and another small model. In our experiments, we specifically use GPT2 as the small model. Note that our approach uses conditional probabilities, whereas Ref does not.

\paragraph{MIN-K\% PROB (MIN-K):}
This method \cite{shi2024detecting} focuses on the minimum token probabilities within a text. It posits that non-member examples are more likely to contain outlier words with high negative log-likelihoods:
\begin{equation}
	\text{MIN-K}(x)= \frac{1}{E} \sum_{x_i \in \text{Min-K\%}(x)} \log \pi_{\theta}(x_i | x_1, ..., x_{i-1}).
\end{equation}
By analyzing these low probability tokens, MIN-K\% PROB provides a distinct method to infer membership, enhancing the diversity of our baseline comparisons.

\section{PROOFS}
\label{app:proofs}

\subsection{Proof of Proposition 1 and 2}
\label{app:proof_prop1}

Here we give the proof of Proposition 1. Proof of Proposition 2 follows from Proposition 1 of \cite{li2023policy}.
First we give an upper bound on the difference between theoretical maximum expected reward and expected DPO reward on the preference dataset i.e. $r(\pi_r^{*}) - r(\pi_{\text{DPO}})$

\begin{align}
	& r(\pi_r^{*}) - r(\pi_{\text{DPO}})  \nonumber \\
	&=  \hat{\EE_x}\hat{\EE}_{y\sim \pi_r^{*}} [r(x,y)] - \hat{\EE_x}\hat{\EE}_{y\sim \pi_{\text{DPO}}} [r(x,y)] \\
	& = \hat{\EE_x}\hat{\EE}_{y\sim \pi_r^{*}} [r(x,y)] - \hat{\EE_x}\hat{\EE}_{y\sim \pi_r^{*}} [\hat{r}_d(x,y)] + \hat{\EE_x}\hat{\EE}_{y\sim \pi_r^{*}} [\hat{r}_d(x,y)] \nonumber \\ 
	& - \hat{\EE_x}\hat{\EE}_{y\sim \pi_{\text{DPO}}} [r(x,y)] + \hat{\EE_x}\hat{\EE}_{y\sim \pi_{\text{DPO}}} [\hat{r}_d(x,y)] - \hat{\EE_x}\hat{\EE}_{y\sim \pi_{\text{DPO}}} [\hat{r}_d(x,y)] \\
	& \leq 2\varepsilon_r + \hat{\EE_x}\hat{\EE}_{y\sim \pi_r^{*}} [\hat{r}_d(x,y)] - \hat{\EE_x}\hat{\EE}_{y\sim \pi_{\text{DPO}}} [\hat{r}_d(x,y)] \qquad \text{(by definition of $\eps_r$)} \\
	& \leq 2\varepsilon_r \qquad \text{(by definition of $\pi_{DPO}$)}
\end{align}


Similarly we give an upper bound on the difference between theoretical maximum expected reward and expected PPO reward on the preference dataset i.e. $r(\pi_r^{*}) - r(\pi_{\text{PPO}})$

\begin{align}
	& r(\pi_r^{*}) - r(\pi_{\text{PPO}})  \nonumber \\
	& =  \hat{\EE_x}\hat{\EE}_{y\sim \pi_r^{*}} [r(x,y)] - \hat{\EE_x}\hat{\EE}_{y\sim \pi_{\text{PPO}}} [r(x,y)] \\ 
	& =   \hat{\EE_x}\hat{\EE}_{y\sim \pi_r^{*}} [r(x,y)] -  \hat{\EE_x}\hat{\EE}_{y\sim \pi_r^{*}} [\hat{r}_p(x,y)] + \hat{\EE_x}\hat{\EE}_{y\sim \pi_r^{*}} [\hat{r}_p(x,y)] \nonumber \\
	& - \hat{\EE_x}\hat{\EE}_{y\sim \pi_{\text{PPO}}} [r(x,y)] + \hat{\EE_x}\hat{\EE}_{y\sim \pi_{\text{PPO}}} [\hat{r}_p(x,y)] - \hat{\EE_x}\hat{\EE}_{y\sim \pi_{\text{PPO}}} [\hat{r}_p(x,y)]\\ 
	& \leq 2\varepsilon_r + \hat{\EE_x}\hat{\EE}_{y\sim \pi_r^{*}} [\hat{r}_p(x,y)] - \hat{\EE_x}\hat{\EE}_{y\sim \pi_{\text{PPO}}} [\hat{r}_p(x,y)] \qquad \text{(by definition of $\eps_r$)} \\
	& \leq 2\varepsilon_r + \hat{\EE_x}\hat{\EE}_{y\sim \pi_r^{*}} [\hat{r}_p(x,y)] - \hat{\EE_x}\EE_{y\sim \pi_r^{*}} [\hat{r}_p(x,y)] + \hat{\EE_x}\EE_{y\sim \pi_r^{*}} [\hat{r}_p(x,y)] \nonumber \\ 
	& -  \hat{\EE_x}\hat{\EE}_{y\sim \pi_{\text{PPO}}} [\hat{r}_p(x,y)] +  \hat{\EE_x}\EE_{y\sim \pi_{\text{PPO}}} [\hat{r}_p(x,y)]  -  \hat{\EE_x}\EE_{y\sim \pi_{\text{PPO}}} [\hat{r}_p(x,y)] \\
	& \leq 2\varepsilon_r + 2\varepsilon_y \qquad \text{(by definition of $\eps_y$ and definition of  $\pi_{PPO}$)}
\end{align}


\subsection{Proof of Proposition \ref{prop:overfitting}}

The loss term $\ell(\theta,z_1)$ in equation \ref{eq:score} refers to the negative reward $-\hat{r}(x_1,y)$ in equations \ref{eq:PPO} and \ref{eq:DPO}. If $\pi_{DPO}$ is indeed the global optimizer, then on the preference dataset $\pi_{DPO}$ is guaranteed to pick the $y \in \{y_w,y_l\}$ such that $-\hat{r}(x_1,y)$ is minimized. If this is not true, then $\pi_{DPO}$ is not the global optimizer. On the other hand, no such condition can be imposed on $\pi_{PPO}$, since the response $y$ is sampled on the conditional distribution of the prompt $x$.  Since $\tau$ is assumed to be a constant in equation \ref{eq:score}, loss term for $\pi_{DPO}$ will be at least as low as that of $\pi_{PPO}$ for $z_1$. Hence, the result follows as $\sigma$ is monotonic.

\subsection{Proof of Lemma \ref{lemma:sablay_modified}}

We assume that $n$ iid points are used for training and $\theta$ is the set of parameters obtained after training. We are interested in inferring the membership $m_1$ of $z_1$. We collect the information related to rest of datapoints in $\ct = \{z_2,\dots,z_n\}$. From theorem 1 of \cite{sablayrolles2019white}, we have that the bayes optimal membership $m_1$ of a sample $z_1$ given $\theta$ is given by 

\begin{align}
	\mathcal{M}\left(\theta, z_1\right) & =\mathbb{P}\left(m_1=1 \mid \theta, z_1\right) \\
	& =\mathbb{E}_{\mathcal{T}}\left[\mathbb{P}\left(m_1=1 \mid \theta, z_1, \mathcal{T}\right)\right]
\end{align}

\begin{align}
	& \mathbb{P}\left(m_1=1 \mid \theta, z_1, \mathcal{T}\right) =\frac{\mathbb{P}\left(\theta \mid m_1=1, z_1, \mathcal{T}\right)  \mathbb{P}\left(m_1=1\right)}{\mathbb{P}\left(\theta \mid z_1, \mathcal{T}\right)} \\
	& =\frac{\alpha}{\alpha+\beta}=\sigma\left(\log \left(\frac{\alpha}{\beta}\right)\right) \\
	where \qquad	& \alpha:=\mathbb{P}\left(\theta \mid m_1=1, z_1, \mathcal{T}\right) \mathbb{P}\left(m_1=1\right) ; 	\beta:=\mathbb{P}\left(\theta \mid m_1=0, z_1, \mathcal{T}\right) \mathbb{P}\left(m_1=0\right)
\end{align}

If we assume a uniform prior over $m_1$, then we have \begin{align}
	\log \left(\frac{\alpha}{\beta}\right)=\log \left(\frac{\mathbb{P}\left(\theta \mid m_1=1, z_1, \mathcal{T}\right)}{\mathbb{P}\left(\theta \mid m_1=0, z_1, \mathcal{T}\right)}\right)
\end{align}

Next, \cite{sablayrolles2019white} defines a posterior $p(\theta)$ based on the likelihood obtained from $\{z_1,\dots,z_n\}$. However, for conveninence in later steps, we define the posterior as follows:

\begin{align}
	p_{\mathcal{T}}(\theta) := \frac{e^{-\frac{1}{T} \sum_{i=1}^n m_i \ell ({\theta}, z_i)}}{\int_t e^{-\frac{1}{T} \sum_{i=1}^n m_i \ell (t, z_i)} \, dt}
\end{align} 

It is straightforward to see that 

\begin{align}
	&\alpha = p_{\ct}(\theta) \\
	&\beta = \frac{e^{\frac{\ell(\theta,z_1)}{T}}p_{\ct}(\theta)}{\int_{t} e^{\frac{\ell(t,z_1)}{T}}p_{\ct}(t) dt}
\end{align}

Thus the bayes optimal membership is given by 

\begin{align}
	\cM(\theta,z_1) = \EE_{\ct}\mbrkt{\sigma\mbrkt{\log\mbrkt{\int_{t} e^{\frac{\ell(t,z_1)}{T}}p_{\ct}(t) dt} - \frac{\ell(\theta,z_1)}{T}}}
\end{align}

\subsection{Proof of Lemma \ref{lemma:ours_overfitting}}
\tdplotsetmaincoords{70}{120}
\begin{figure}[H]
	\centering
	\begin{tikzpicture}[scale=5,tdplot_main_coords]
		\coordinate (A) at (0,0,0);
		\coordinate (B) at (1,0,0);
		\coordinate (C) at (0,1,0);
		\coordinate (D) at (0,0,1);
		
		\coordinate (Ap) at (0,0,1-0.35);
		\coordinate (Bp) at (0.35,0,1-0.35);
		\coordinate (Cp) at (0,0.35,1-0.35);
		\coordinate (Dp) at (0, 0, 1);
		
		\draw[thick] (A) -- (B) -- (C) -- cycle;
		\draw[thick] (A) -- (D) -- (C);
		\draw[thick] (A) -- (B) -- (D);
		\draw[thick] (B) -- (C) -- (D);
		
		\draw[dashed,thick] (Ap) -- (Bp) -- (Cp) -- cycle;
		\draw[dashed,thick] (Ap) -- (Dp) -- (Cp);
		\draw[dashed,thick] (Ap) -- (Bp) -- (Dp);
		\draw[dashed,thick] (Bp) -- (Cp) -- (Dp);
		
		\node[above left] at (A) {$A$};
		\node[above right] at (B) {$B$};
		\node[below right] at (C) {$C$};
		\node[below left] at (D) {$D$};
		
		\node[above left] at (Ap) {$A'$};
		\node[above right] at (Bp) {$B'$};
		\node[below right] at (Cp) {$C'$};

		\coordinate (Aoffset) at (0.15,0.15,0.15);
		\coordinate  (Apoffset) at (0.15,0.15,0.80);
		\draw[decorate,decoration={brace,mirror,amplitude=10pt}] (A) -- (Ap) node[midway,above,yshift=0.0cm,xshift=0.5cm,font=\Large] {$\eps$};
		
	\end{tikzpicture}
	\captionof{figure}{The standard 2-dimensional simplex forms a standard 3-dimensional tetrahedron with the origin $A$ whose volume is given by $\frac{1}{3!}$. The smaller tetrahedron $A'B'C'D$ has a volume $\frac{(1-\eps)^3}{3!}$}
	\label{fig:geometric_intuition}
\end{figure}

Since $L_n \leq \eta$, we have $\sum_{i=1}^{n} \ell(\theta,z_i) < n\eta$, with each individual term in the summation $\geq 0$. Let $\ell_i = \ell(\theta,z_i)$ and since each $z_i$ is an iid sample, the space of $\{\ell_i\}_{i=1}^n$ forms a $n$-dimensional standard tetrahedron $\cv$ of volume $\frac{(n\eta)^n}{n!}$. Within this tetrahedron $\cv$, the space where $\{\ell_1 > \eps\}$ forms a similar tetrahedron of volume maximum $\frac{(n\eta - \eps)^n}{n!}$. For a geometric intuition, refer to Figure \ref{fig:geometric_intuition}. Thus, the probability of the event $\{\ell_1 < \eps\}$ is given by 

\begin{align}
	\mP\left[\{\ell_1 < \eps\}\right] &=  1 - 	\mP\left[\{\ell_1 \geq \eps\}\right]      \\
	&\geq 1 -  \frac{(n\eta - \eps)^n}{(n\eta)^n} \\
	&\geq \mbrkt{1-\mbrkt{1-\frac{1}{n\alpha}}^{n}} \qquad \text{(by substituting $\eta = \eps\alpha$)} 
\end{align}

Below we comparison of the lower bounds given here and that of \cite{aubinais2023fundamental}.
\begin{figure}[htbp]
	\centering
	\includegraphics[width=0.5\textwidth]{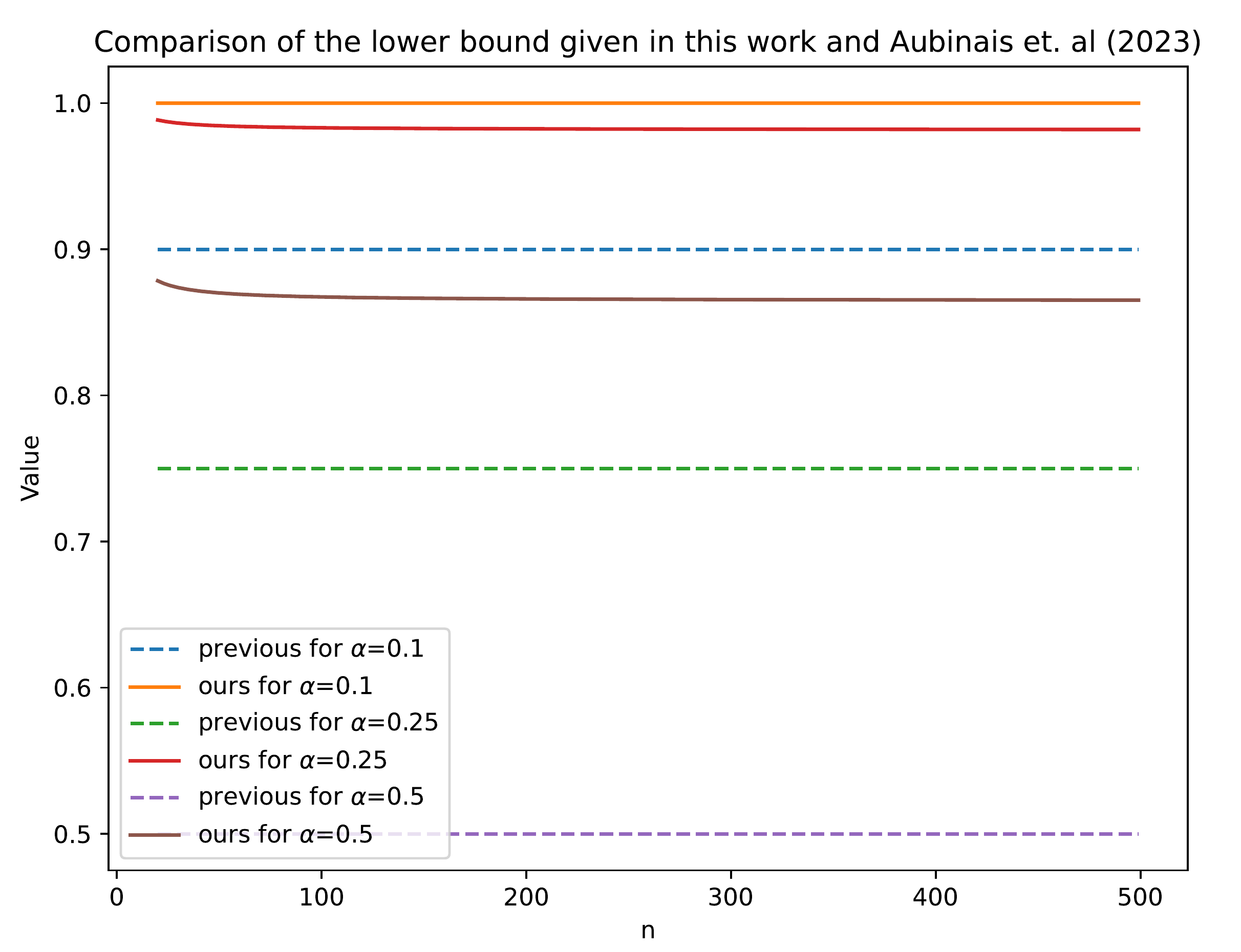}
	\caption{The lower bound given here is much tighter compared to that of \cite{aubinais2023fundamental}}
	\label{fig:comparison_lowerbounds}
\end{figure}

\subsection{Proof of Theorem \ref{thm:lowerbound}}

%
%
%
%

\begin{proof}{Proof of theorem \ref{thm:lowerbound}}
	
	Let $B_\eps := \{\theta : \ell(\theta,z_1) \leq \eps \}$. Then, 
	
	\begin{align}
		& s(z_1,{\theta}, p_{\mathcal{T}})  =  \log \left( \int_{t} e^{ \frac{1}{T}\ell(t, z_1) } p_{\mathcal{T}}(t) dt \right) - \frac{\ell(\theta,z_1)}{T} \qquad \text{(ignoring $1/T$ for the time being)}\\
		& = \log \mbrkt{ \int_{B_{\eps}}  e^{ \frac{1}{T}\ell(t, z_1) } p_{\mathcal{T}}(t) dt + \int_{B_{\eps}^{C}}  e^{ \frac{1}{T}\ell(t, z_1) } p_{\mathcal{T}}(t) dt} -  \frac{\ell(\theta,z_1)}{T} \\
		& \geq \log \mbrkt{\mbrkt{1-\mbrkt{1-\frac{1}{n\alpha}}^{n}} + \int_{B_{\eps}^{C}}  e^{ \frac{1}{T}\ell(t, z_1) } p_{\mathcal{T}}(t) dt} -  \frac{\ell(\theta,z_1)}{T} \quad \text{(substituting $\ell = 0$ for all $\theta \in B_\eps$ and from Lemma \ref{lemma:ours_overfitting})}\\
		& \geq \log \mbrkt{1-\mbrkt{1-\frac{1}{n\alpha}}^{n}} -  \frac{\ell(\theta,z_1)}{T} \quad \text{(dropping the second integral term which $\geq 0$ almost surely)}
	\end{align}
\end{proof}

\section{IMPLEMENTATION DETAILS}
\label{app:implementation_details}
We mainly refer to the TRL\footnote{\url{https://huggingface.co/docs/trl/en/index}} package for implementation.

\textbf{LoRA Setting.} For all experiments, we share the same LoRA setting below, using the PEFT\footnote{\url{https://huggingface.co/docs/peft/index}} package: \texttt{lora\_alpha} 32, \texttt{lora\_dropout} 0.05, \texttt{lora\_r} 16, and no bias term.

\textbf{Quantization Setting.} For all experiments, we use the BitsAndBytes\footnote{\url{https://huggingface.co/docs/bitsandbytes/index}} package for 4-bit quantization.

\textbf{SFT Setting.} The settings for SFT are detailed below. We utilized the "train/rl" split of the stack-exchange-paired dataset, selecting 80,000 data points for the fine-tuning process, same data is used for PPO and DPO training. The prompt and only the preferred response are concatenated as input. The specific training parameters are:\\
- \textbf{Training Epochs:} 2.0\\
- \textbf{Learning Rate:} 8e-5\\
- \textbf{Batch Size (Training):} 4\\
- \textbf{Batch Size (Evaluation):} 2\\
- \textbf{Gradient Accumulation Steps:} 4\\
- \textbf{Learning Rate Scheduler:} cosine\\
- \textbf{Warmup Steps:} 100\\
- \textbf{Weight Decay:} 0.05\\
- \textbf{Optimizer:} paged\_adamw\_32bit\\
- \textbf{Mixed Precision Training:} fp16\\

\textbf{PPO Setting.} The settings for PPO are detailed below. We filter out data points with maximum length constraints. We also limit the maximum length of the generated response. The specific training parameters are:
\\
- \textbf{Batch Size:} 16\\
- \textbf{Mini Batch Size:} 4\\
- \textbf{Gradient Accumulation Steps:} 4\\
- \textbf{PPO Epochs:} 6\\
- \textbf{Learning Rate:} 5.4e-5\\
- \textbf{KL Coefficient:} 0.1\\
- \textbf{Adaptive KL Control:} True\\
- \textbf{Target KL:} 5.0\\
- \textbf{Horizon:} 4000\\
- \textbf{Training Epochs:} 4\\
- \textbf{Maximum Output Length:} 128\\
- \textbf{Maximum Prompt Length:} 256\\
- \textbf{Maximum Sequence Length:} 1024\\

\textbf{DPO Setting.} The settings for DPO training are detailed below. The specific training parameters are:
\\
- \textbf{Batch Size (Training):} 8\\
- \textbf{Batch Size (Evaluation):} 2\\
- \textbf{Gradient Accumulation Steps:} 2\\
- \textbf{Training Epochs:} 3.0\\
- \textbf{Learning Rate:} 5e-4\\
- \textbf{Warmup Steps:} 100\\
- \textbf{Maximum Sequence Length:} 1024\\
- \textbf{Maximum Prompt Length:} 256\\
- \textbf{Optimizer Type:} paged\_adamw\_32bit\\
- \textbf{Beta:} 0.4\\

\section{ADDITIONAL EXPERIMENTAL RESULTS}

\subsection{Results of Experiments on GPT2 Series}
\label{app:GPT2_exp}

In Table \ref{tab:Mink-SE-GPT2}, we give the AUROC scores for other frameworks with GPT2 series on SE dataset. Compared to the optimistic PREMIA, we find that the traditional MIA methods are not that effective when it comes to MIA on smaller models such as GPT2-series. Nevertheless, here too, we consistently notice that susceptibility of DPO is higher compared to PPO. 

\begin{table}[h]
	\centering
	\tiny
	\caption{AUROC scores comparing different MIA methods on GPT2-series on SE dataset}
	\label{tab:Mink-SE-GPT2}
	\begin{tblr}{
			column{1} = {c},
			cell{1}{3} = {c=2}{c},
			cell{1}{5} = {c=2}{c},
			cell{2}{2} = {c},
			cell{2}{3} = {c},
			cell{2}{4} = {c},
			cell{3}{2} = {c},
			cell{3}{3} = {c},
			cell{3}{4} = {c},
			cell{4}{1} = {r=5}{},
			cell{4}{2} = {c},
			cell{4}{3} = {c},
			cell{4}{4} = {c},
			cell{5}{2} = {c},
			cell{5}{3} = {c},
			cell{5}{4} = {c},
			cell{6}{2} = {c},
			cell{6}{3} = {c},
			cell{6}{4} = {c},
			cell{7}{2} = {c},
			cell{7}{3} = {c},
			cell{7}{4} = {c},
			cell{8}{2} = {c},
			cell{8}{3} = {c},
			cell{8}{4} = {c},
			cell{9}{1} = {r=5}{},
			cell{9}{2} = {c},
			cell{9}{3} = {c},
			cell{9}{4} = {c},
			cell{10}{2} = {c},
			cell{10}{3} = {c},
			cell{10}{4} = {c},
			cell{11}{2} = {c},
			cell{11}{3} = {c},
			cell{11}{4} = {c},
			cell{12}{2} = {c},
			cell{12}{3} = {c},
			cell{12}{4} = {c},
			cell{13}{2} = {c},
			cell{13}{3} = {c},
			cell{13}{4} = {c},
			cell{14}{1} = {r=5}{},
			cell{14}{2} = {c},
			cell{14}{3} = {c},
			cell{14}{4} = {c},
			cell{15}{2} = {c},
			cell{15}{3} = {c},
			cell{15}{4} = {c},
			cell{16}{2} = {c},
			cell{16}{3} = {c},
			cell{16}{4} = {c},
			cell{17}{2} = {c},
			cell{17}{3} = {c},
			cell{17}{4} = {c},
			cell{18}{2} = {c},
			cell{18}{3} = {c},
			cell{18}{4} = {c},
			cell{19}{1} = {r=5}{},
			cell{19}{2} = {c},
			cell{19}{3} = {c},
			cell{19}{4} = {c},
			cell{20}{2} = {c},
			cell{20}{3} = {c},
			cell{20}{4} = {c},
			cell{21}{2} = {c},
			cell{21}{3} = {c},
			cell{21}{4} = {c},
			cell{22}{2} = {c},
			cell{22}{3} = {c},
			cell{22}{4} = {c},
			cell{23}{2} = {c},
			cell{23}{3} = {c},
			cell{23}{4} = {c},
			vlines,
			hline{1,24} = {-}{0.08em},
			hline{2-4,9,14,19} = {-}{},
		}
		&                  & Chosen   &          & Rejected &          \\
		& Framework            & DPO & PPO & DPO & PPO \\
		Modelname          &                  &          &          &          &          \\
		gpt2        & MIN-K          & 0.545    & 0.526    & 0.523    & 0.524    \\
		& PPL              & 0.538    & 0.528    & 0.519    & 0.524    \\
		& Ref       & 0.556    & 0.521    & 0.542    & 0.521    \\
		& Lowercase & 0.559    & 0.519    & 0.523    & 0.532    \\
		& Zlib         & 0.520    & 0.518    & 0.528    & 0.532    \\
		gpt2-medium & MIN-K          & 0.516    & 0.515    & 0.522    & 0.530    \\
		& PPL              & 0.512    & 0.516    & 0.517    & 0.529    \\
		& Ref      & 0.526    & 0.516    & 0.543    & 0.508    \\
		& Lowercase & 0.543    & 0.514    & 0.528    & 0.521    \\
		& Zlib        & 0.528    & 0.525    & 0.514    & 0.520    \\
		gpt2-large  & MIN-K          & 0.525    & 0.517    & 0.537    & 0.518    \\
		& PPL         & 0.521    & 0.518    & 0.531    & 0.520    \\
		& Ref       & 0.528    & 0.527    & 0.537    & 0.528    \\
		& Lowercase & 0.566    & 0.529    & 0.538    & 0.528    \\
		& Zlib        & 0.519    & 0.527    & 0.524    & 0.512    \\
		gpt2-xl     & MIN-K         & 0.524    & 0.521    & 0.536    & 0.514    \\
		& PPL           & 0.519    & 0.518    & 0.533    & 0.509    \\
		& Ref      & 0.528    & 0.511    & 0.549    & 0.518    \\
		& Lowercase & 0.566    & 0.506    & 0.538    & 0.512    \\
		& Zlib        & 0.520    & 0.525    & 0.520    & 0.528    
	\end{tblr}
\end{table}

\subsection{Impact of Response Length on MIA Effectiveness}
\begin{figure}[h]
\centering
\includegraphics[width=0.75\textwidth]{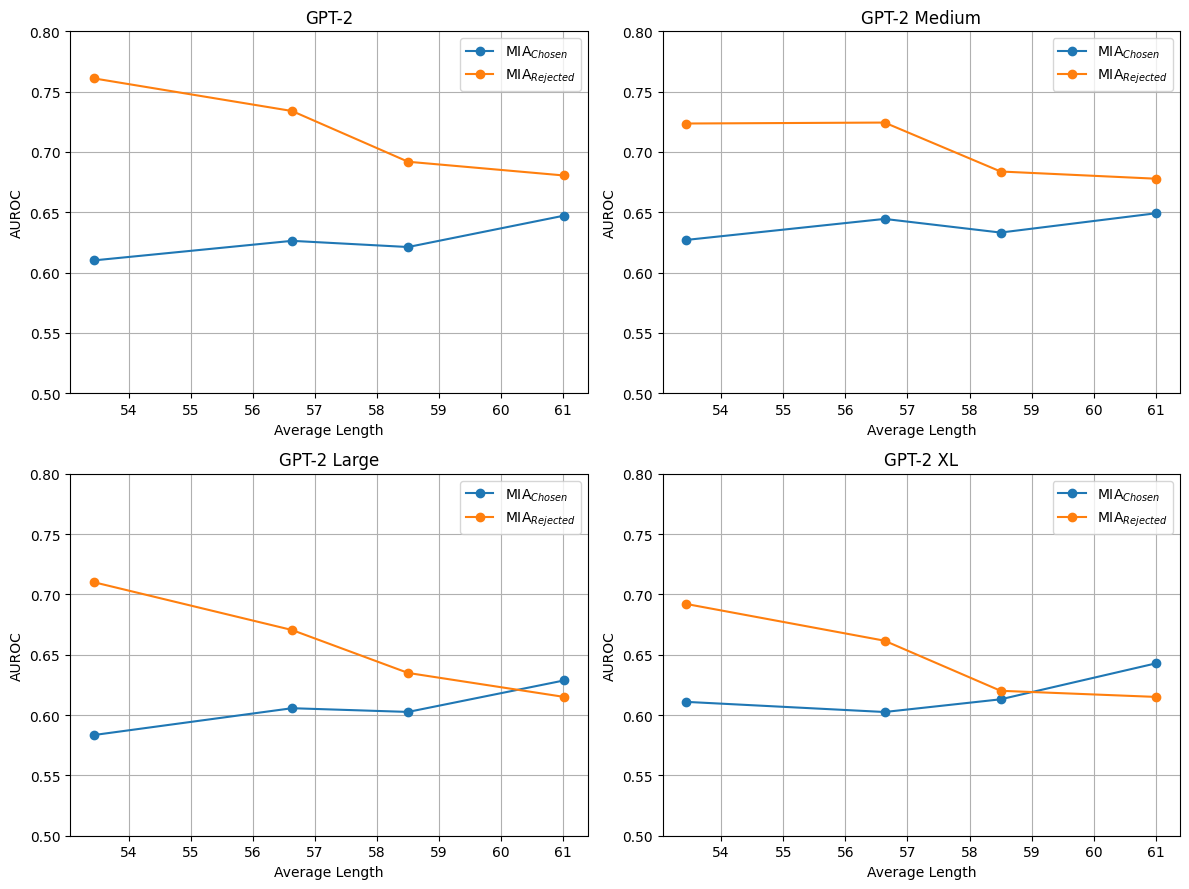}
\caption{AUROC vs Average Length for GPT-2 Models}
\label{fig:len_vs_roc}
\end{figure}

{In this experiment, we look the effect of length of examples used in preference alignment and their corresponding vulnerability in terms of AUC-ROC of PREMIA-SFT.}
Figure~\ref{fig:len_vs_roc} shows the MIA AUROC results for the GPT-2 family of models on the IMDB dataset. As can be seen from the figure, for "Chosen" responses, the longer the response, the more susceptible it is to MIA, while for "Rejected" responses, the opposite is true.
\textcolor{black}{Note that in general, increasing sequence length leads to higher AUC as longer texts tend to have more memorized information by the model \citep{shi2024detecting}. However, the 'Chosen' response is used during both SFT and DPO alignment whereas the 'Rejected' response is used only during DPO, thus the difference in their trends when measured using PREMIA-SFT. 
	 Since in our PREMIA-SFT framework, we divide by the conditional probability of the SFT model, the higher probability of the longer texts in DPO gets negated by the higher probability in SFT model. }

\subsection{AUROC plot for pair detection for IMDB dataset}
\label{app:imdb_pair_plot}
\begin{minipage}[th]{.85\linewidth}
	\centering
	\captionsetup{width=\linewidth}
	\captionof{figure}{AUROC scores for $\text{MIA}_\text{Pair}$ detection for Mistral-7B, Open-llama-3b, and Open-llama-7b models for IMDB dataset. As mentioned in \S \ref{sec:MIA_Methodology_Effectiveness}, for easier tasks such as the IMDB sentiment alignment, both PPO and DPO are robust to MIA.}
	\includegraphics[width=\linewidth]{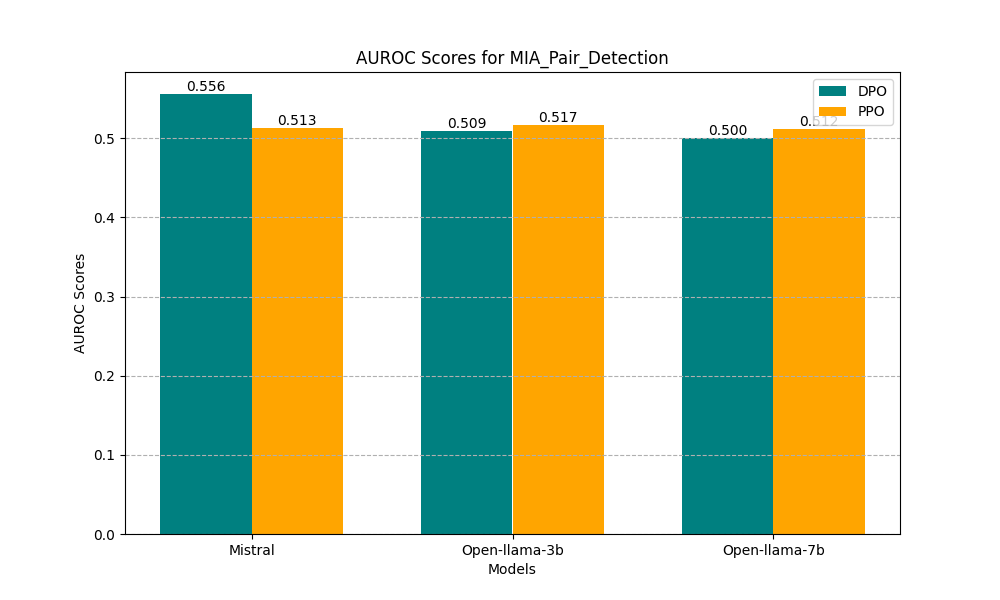}
	
	\label{fig:MIA_Pair_BarPlot_imdb}
	
\end{minipage}

\subsection{Anecdotal responses by Mistral-7B models on SE dataset}
\label{app:anecdotal_responses}

\begin{tabularx}{\textwidth}{@{} l X @{}}
	\toprule
Question & What are some grilling tools that a beginner should have when purchasing a new grill? Additionally, are there tools that are useful for the more advanced outdoor cook? \\
	& \\
	\midrule
	Chosen & Tool-wise: spatula, tongs, grill brush (for cleaning) and a basting brush. If you're using extremely hot fire or high flames there are grill-safe mitts you can buy. \\
	\midrule
Rejected &  I have a wide, flat knife that can be used as a spatula; I find it really useful. Everything else has pretty much been mentioned, but I would suggest a pair of tongs with decent teeth on them. It's very disheartening when that burger or steak slips out of your grip and into the coals or onto the ground. \\
	\midrule
	Base & 1. A good set of tongs.
	
	2. A good set of spatulas.
	
	3. A good set of skewers.
	
	4. A good set of grill brushes.
	
	5. A good set of grill gloves.
	
	6. A good set of grill tools.
	
	7. A good set of grill accessories.
	
	8. A good set of grill accessories.
	
	9. A good set of grill accessories. \\
	\midrule
SFT & 1. A good set of tongs.

2. A good set of spatulas.

3. A good set of skewers.

4. A good set of grill brushes.

5. A good set of grill gloves.

6. A good set of grill tools.

7. A good set of grill baskets.

8. A good set of grill baskets.

9. A good set of grill baskets.

10. A good set of grill baskets.

11. A good set of grill baskets.
 \\
	\midrule
	DPO & **Beginner:**
	
	* **Tongs:** These are the most important tool. You'll use them for everything from flipping burgers to moving hot coals. Get a pair that are comfortable to hold and have a good grip.
	* **Spatula:** This is for flipping things like pancakes and eggs. You'll also use it to scrape the grill grates clean.
	* **Grill brush:** This is for cleaning the grill grates. You'll want to get one with stiff bristles.
	* **Thermometer:** This is for checking the temperature of the grill. You'll want to get a digital one that has a probe.
	
	**Advanced:**
	
	* **Skewers:** These are for making kebabs.
	* **Basting brush:** This is for brushing sauces on meat.
	* **Grill basket:** This is for cooking vegetables and other things that are small and fall through the grates.
	* **Grill mat:** This is for cooking things that are delicate and fall through the grates.
	* **Grill gloves:** These are for protecting your hands from the heat.
	* **Grill light:** This is for seeing what you're doing at night.
	
	**Both:**
	
	* **Grill cleaner:** This is for cleaning the grill grates.
	* **Grill cover:** This is for protecting the grill from the elements. \\
	\midrule
	PPO & 1. A good set of tongs
	
	2. A good set of grill brushes
	
	3. A good set of grill tools (spatula, fork, etc.)
	
	4. A good set of grill gloves
	
	5. A good set of grill mats
	
	6. A good set of grill baskets \\
	\midrule
	\midrule

	\end{tabularx}

\begin{tabularx}{\textwidth}{@{} l X @{}}
	\toprule
Question & Does `dbms\_output.put\_line` decrease the performance in `plsql` code? \\
& \\
\midrule
Chosen & Yes, it's another piece of code that needs to be executed, but unless the output is actually turned on, I think the overhead is quite minimal. 		Here's an AskTom question with more details: Is there a performance impact for dbms\_output.put\_line statements left in packages? \url{	
	http://asktom.oracle.com}
\\
\midrule
Rejected &  You can look into [conditional compilation](http://www.oracle-base.com/articles/10g/ConditionalCompilation\_10gR2.php) so that the DBMS\_OUTPUT.PUT\_LINE are only in the pre-parsed code if the procedure is compiled with the appropriate option. \\
\midrule
Base & No, it does not.

`dbms\_output` is a package that is used to display the output of a PL/SQL block. It is not a performance issue.

`dbms\_output` is a package that is used to display the output of a PL/SQL block. It is not a performance issue.

The `dbms\_output` package is used to display the output of a PL/SQL block. It is not a performance issue.

The `dbms\_output` package is used to display the output of a PL/SQL block. It is not a performance issue.

The `dbms\_output` package is used to display the output of a PL/SQL block. It is not a performance issue.
 \\
\midrule
SFT & 1. `dbms\_output` is a package that is used to display output from PL/SQL blocks.
2. `dbms\_output\.put\_line` is a procedure that is used to display a line of output from PL/SQL blocks.
3. `dbms\_output` is not a function, it is a package.
4. `dbms\_output.put\_line` is not a function, it is a procedure.
5. `dbms\_output` is not a variable, it is a package.
6. `dbms\_output.put\_line` is not a variable, it is a procedure.
7. `dbms\_output` is not a constant, it is a package.
8. `dbms\_output.put\_line` is not a constant, it is a procedure.
\\
\midrule
DPO &  **Yes, it does.**

`dbms\_output` is meant for debugging only. It is not meant for production code.

From the documentation:
> **dbms\_output** 
The DBMS\_OUTPUT package provides a mechanism for

\\
\midrule
PPO & Yes, it does.

`dbms\_output` is a package that is used to display output from a PL/SQL block. It is not a package that is used to display output from a SQL statement.

The `dbms\_output` package is a package that is used to display output from a PL/SQL block. \\
\midrule
\toprule

\end{tabularx}


\end{document}